\documentclass[letterpaper]{article} 
\usepackage{aaai2027}  
\usepackage[hyphens]{url}  
\usepackage{graphicx} 
\usepackage{subcaption}
\def\UrlFont{\rm}  
\usepackage{natbib}  
\usepackage{caption} 
\usepackage{amsmath}
\usepackage{amssymb}
\usepackage{algorithm}
\usepackage{algorithmic}
\newcommand{\bestscore}[1]{\textbf{#1}}
\newcommand{\secondscore}[1]{\underline{#1}}
\usepackage{newfloat}
\usepackage{listings}
\DeclareCaptionStyle{ruled}{labelfont=normalfont,labelsep=colon,strut=off} 
\floatstyle{ruled}
\newfloat{listing}{tb}{lst}{}
\floatname{listing}{Listing}

\usepackage{booktabs}

\title{Aligning Multi-Trajectory Supervision with Policy Optimization for VLA Driving}

\author{
    Tian Zhang\textsuperscript{\rm 1,\rm 2},
    Zhuo Huang\textsuperscript{\rm 2},
    Hongrui Ye\textsuperscript{\rm 1,\rm 2},
    Yu Wu\textsuperscript{\rm 2},\\
    Zengmao Wang\textsuperscript{\rm 1},
    Kaixuan Zhou\textsuperscript{\rm 2}\textsuperscript{\ensuremath{\dagger}}
}
\affiliations{
    \textsuperscript{\rm 1}School of Computer Science, Wuhan University, China\\
    \textsuperscript{\rm 2}Dongfeng Research \& Development Institute\\
    \{tianzhang, yehonngrui, wangzengmao\}@whu.edu.cn\\
    \{huangzhuo, dfrd-wuyu, tc-zhoukaixuan\}@dfmc.com.cn
}

\begin{document}

\maketitle

\begingroup
\renewcommand{\thefootnote}{\fnsymbol{footnote}}
\footnotetext[2]{Project leader. Work done when Tian Zhang was an intern at
Dongfeng Research \& Development Institute.}
\endgroup

\begin{abstract}
Vision-language-action (VLA) driving methods increasingly combine multi-trajectory imitation learning with group-relative policy optimization (GRPO), making trajectory selection critical to final performance. However, some high-scoring trajectories that improve imitation can degrade subsequent GRPO by inducing advantage estimates misaligned with the current policy’s feasible behavior distribution, driving updates away from safe and compliant behaviors. To address this, we propose a novel framework that aligns multi-trajectory supervision with policy optimization. To address the policy gradient bias induced by infeasible noisy trajectories outside the feasible region, augmented trajectories are constrained to a neighboring manifold of the ground-truth feasible region, and a Pareto-optimality criterion is adopted in place of the conventional aggregate score, retaining only non-dominated candidates and thereby filtering out conflicting samples at the source. To ensure that expanded trajectory supervision is effectively absorbed during policy optimization, we introduce two complementary mechanisms: feasibility-first advantage assignment and dynamic distillation. The former adapts Pareto credit to the feasibility composition of each rollout group and guides fully infeasible groups toward safe references. The latter updates teacher trajectories across refinement rounds to continually transfer useful supervision. Together, they progressively translate the benefits of expanded supervision into policy improvement. On NAVSIM v1 and v2, our method achieves 91.4 PDMS and 89.1 EPDMS, respectively, under single-trajectory inference, and recovers 440 of 658 initially failed scenes, 11.1\% higher than the original GRPO baseline.

\end{abstract}


\section{Introduction}
\label{sec:introduction}

\begin{figure}[!t]
    \centering
    \begin{subfigure}[t]{0.98\columnwidth}
        \centering
        \includegraphics[width=\linewidth]{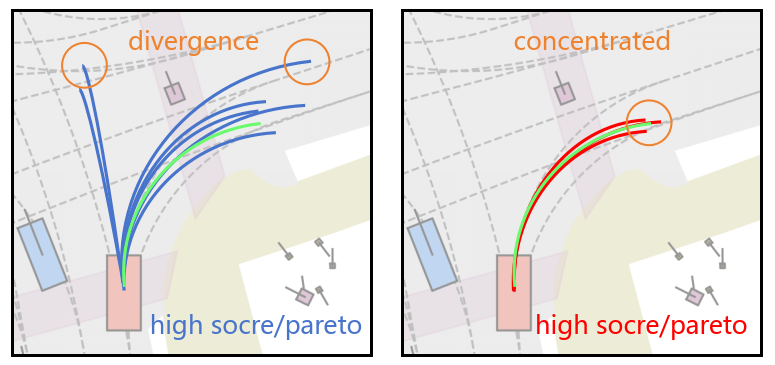}
        \caption{Conventional and policy-compatible trajectory selection.}
        \label{fig:subfig_a}
    \end{subfigure}
    \smallskip
    \begin{subfigure}[t]{0.98\columnwidth}
        \centering
        \includegraphics[width=\linewidth]{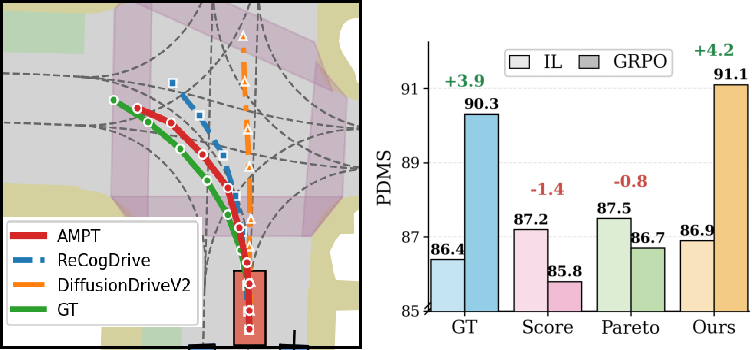}
        \caption{Representative trajectories and IL-to-GRPO performance.}
        \label{fig:subfig_b}
    \end{subfigure}
    \caption{\textbf{Motivation for policy-compatible trajectory selection.}
    (a) Conventional selection based on score, Pareto filtering, or geometric diversity may retain high-quality candidates far from trajectories sampled by the frozen imitation policy, whereas our method selects candidates that are also compatible with the policy distribution.
    (b) Although score-based and Pareto-based supervision improves the imitation-learning checkpoints over GT-only training, both policies deteriorate after GRPO. The comparison shows that trajectory quality alone does not determine its value for downstream policy optimization.}
    \label{fig:trajectory}
\end{figure}

Vision-language-action (VLA) models are increasingly used in end-to-end driving to integrate visual scene understanding, high-level reasoning, and trajectory generation within a unified policy~\cite{tian2024drivevlm,hwang2024emma,wang2025omnidrive,li2025recogdrive}. Recent methods commonly learn an initial policy from logged demonstrations and then refine it through group-relative policy optimization (GRPO) with planning-oriented rewards~\cite{li2025recogdrive,jiang2025alphadrive,zou2025diffusiondrivev2}. Because a logged future represents only one valid response to a driving scene, multi-trajectory imitation is used to broaden the learned behavior distribution before post-training~\cite{liao2025diffusiondrive,chen2026curiousvla,ang2026clover}. Additional trajectories are typically selected according to aggregate planning scores, physical validity, or geometric diversity. These criteria characterize the quality of a candidate as an isolated planning outcome, but do not indicate whether learning it produces a policy that can be improved effectively by subsequent GRPO.

We identify a supervision--optimization mismatch in this training paradigm.
\textbf{A high-scoring trajectory is not necessarily a useful supervision target because its value also depends on its compatibility with the policy being trained.} As illustrated in Fig.~\ref{fig:trajectory}(a), selection based on evaluator score, Pareto filtering, or geometric diversity may retain candidates far from trajectories sampled by the frozen imitation policy.
Fig.~\ref{fig:trajectory}(b) shows the resulting consequence: score-based and Pareto-based supervision improves the imitation checkpoints to 87.2 and 87.5 PDMS, but the corresponding policies deteriorate to 85.8 and 86.7 after GRPO. In contrast, GT-only supervision and our policy-compatible selection continue to improve during post-training, reaching 90.3 and 91.1 PDMS.
Distant targets can induce large shifts in the learned trajectory distribution, while their high nominal scores do not guarantee feasibility under prediction or sampling errors. Effective trajectory expansion should therefore consider both planning quality and compatibility with the current policy. This naturally motivates us consider that how can multi-trajectory supervision be constructed to support downstream policy optimization?

The same alignment problem also arises in GRPO credit assignment. Standard GRPO ranks rollouts using a scalar planning reward and increases the likelihood of samples that perform well relative to the group~\cite{li2025recogdrive,zou2025diffusiondrivev2}. Driving evaluation, however, combines safety and compliance requirements with objectives such as progress, time-to-collision margin, and comfort. A scalar score can conceal this structure: progress gains may compensate for safety regressions, and a rollout may receive positive credit even when another feasible sample performs no worse across all relevant objectives. The problem is more severe when every rollout in a group is infeasible, because relative ranking provides no reliable direction toward safe behavior. Effective policy optimization should therefore establish or recover feasibility before improving driving efficiency within the feasible set.

Alignment must also be maintained as the policy evolves. On-policy GRPO may not fully absorb useful behaviors available in trajectories generated by different policy seeds, structured trajectory expanders, or policies specialized for safety and progress. At the same time, the value of these candidates changes with the policy: a trajectory that improves an earlier checkpoint may become redundant after an update, while a previously unsuitable candidate may become useful. Reusing a fixed teacher set can therefore introduce outdated supervision, whereas directly imitating distant teachers may disturb behaviors already acquired by the current policy. Further refinement requires candidate trajectories to be reassessed against the latest policy and transferred through conservative updates.

We propose Aligned Multi-Trajectory Policy Training (AMPT), a unified framework that aligns multi-trajectory supervision, rollout credit assignment, and iterative policy refinement. First, policy-compatible multi-trajectory supervision (PC-MTS) retains candidates on the scene-wise Pareto front only when they are close to rollouts from a frozen GT-only policy and remain feasible under small trajectory deviations. Fine-tuning a copy of the frozen policy on the logged and accepted trajectories produces a policy-compatible initialization for GRPO, avoiding the degradation caused by individually strong but incompatible supervision. Second, feasibility-first Pareto GRPO (FF-PGRPO) penalizes unsafe rollouts and grants positive credit only to feasible rollouts that do not regress from a scene-specific reference and lie on the group Pareto front. When no feasible rollout is sampled, a safe reference provides recovery supervision, ensuring that driving efficiency is improved only after feasibility has been established. Third, adaptive Pareto refinement (APR) reassesses a multi-source teacher pool against the latest policy, converts selected teachers into re-evaluated local targets, and transfers them through advantage-weighted distillation with policy retention. Rebuilding the teacher set after each round removes outdated supervision while preserving previously learned behavior. Under single-trajectory inference, AMPT achieves 91.4 PDMS and 89.1 EPDMS on NAVSIM v1 and v2, respectively, and recovers 440 of 658 initially failed scenes, 73 more than the original GRPO baseline.

The main contributions can be summarized as:

    


\begin{itemize}
    \item A structural supervision--optimization mismatch is identified
    in multi-trajectory VLA driving: trajectory sets that improve
    imitation-stage planning can reduce the availability of feasible,
    high-quality rollouts during subsequent GRPO. This shows that
    individual trajectory scores are insufficient for assessing
    downstream optimization value.

    \item Policy-compatible multi-trajectory supervision (PC-MTS) is
    introduced to construct an optimization-oriented trajectory set.
    High-scoring candidates are screened by component-wise Pareto
    optimality and retained only when they remain close to rollouts from
    a frozen GT-only policy, filtering policy-mismatched supervision
    before fine-tuning.

    \item Two complementary mechanisms are developed to exploit expanded
    supervision. Feasibility-first Pareto GRPO (FF-PGRPO) adapts credit
    assignment to rollout-group feasibility and guides fully infeasible
    groups toward safe references. Adaptive Pareto-guided policy refinement (APR) selects teacher trajectories that improve the
    current policy without degrading protected metrics and transfers
    them through advantage-weighted distillation with policy retention.
\end{itemize}

\begin{figure*}[t]
    \centering
    \IfFileExists{Figures/framework.png}{%
        \includegraphics[
            width=0.98\textwidth,
            keepaspectratio
        ]{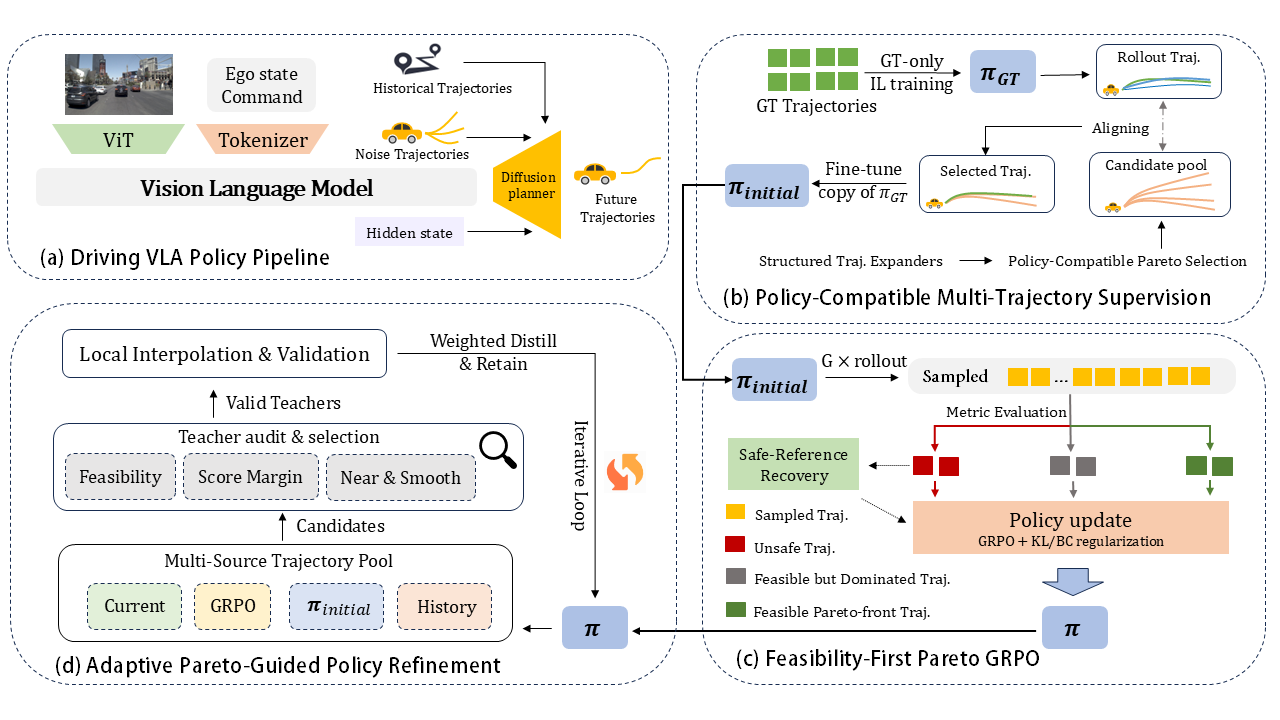}%
    }{%
        \fbox{\rule{0pt}{1.45in}\rule{0.94\textwidth}{0pt}}%
    }
    \caption{\textbf{Overview of AMPT.}(a) A driveing VLA diffusion policy predicts future trajectories.(b) PC-MTS selects high-quality candidates that are compatible with the frozen GT-only policy $\pi_{GT}$ and locally feasible, yielding $\pi_{\mathrm{initial}}$. (c) FF-PGRPO assigns rollout credit according to feasibility, reference-relative performance, and Pareto comparison, with safe-reference recovery for fully infeasible groups. (d) APR repeatedly validates and distills policy-relative improvements from a multi-source trajectory pool.}
    \label{fig:framework}
\end{figure*}

\section{Related Work} \paragraph{Multi-Trajectory Supervision for Driving Policies.} VLA driving models connect scene understanding and high-level reasoning with trajectory generation through unified multimodal policies or language-conditioned action decoders~\cite{tian2024drivevlm,wang2025omnidrive,hwang2024emma,zhou2025opendrivevla}. Diffusion-based planners represent multiple plausible futures using motion anchors, proposal sets, learned scoring, or simulator-generated alternatives~\cite{chi2023diffusionpolicy,liao2025diffusiondrive,wang2025hmad,wang2026drivejepa}. Although these methods improve trajectory coverage and proposal quality, additional targets are still selected mainly by consistency with demonstrations, evaluator scores, or geometric diversity. Curious-VLA explicitly connects the imitation distribution to subsequent reinforcement learning and shows that an overly concentrated policy limits rollout exploration~\cite{chen2026curiousvla}. Whereas Curious-VLA focuses on broadening the policy distribution, we study which additional trajectories provide useful supervision for downstream optimization. Our selection therefore considers planning quality together with compatibility with rollouts sampled by the frozen imitation policy and local feasibility. \paragraph{Reinforcement Post-Training and Policy Refinement.} GRPO has been applied to both driving reasoning and continuous trajectory policies~\cite{li2025recogdrive,jiang2025alphadrive,tang2025planr1}. DiffusionDriveV2 structures comparisons within and across motion anchors~\cite{zou2025diffusiondrivev2}, while HAD and EvaDrive expose metric-specific or multi-objective feedback during policy optimization~\cite{yao2026had,jiao2025evadrive}. Unlike reward decomposition or weighted scalarization, our method treats safety and compliance as prerequisites and applies Pareto comparison only among feasible rollouts to determine eligibility for positive credit. Our final stage is related to conservative policy improvement and trajectory distillation. Support-constrained, baseline-aware, and advantage-weighted methods limit policy shift while preserving prior behavior~\cite{kumar2019bear,laroche2019spibb,nair2020awac}. CLOVER constructs evaluator-filtered pseudo-experts and refines a generator toward scorer-selected Pareto targets, retaining proposal ranking at inference~\cite{ang2026clover}. We instead treat teacher validity as policy-dependent: candidates are reassessed after every round, converted into locally validated targets, and distilled with behavior retention. The resulting policy predicts a single trajectory without a test-time scorer or candidate reranking.

\section{Preliminaries}
\label{sec:preliminaries}

\paragraph{Driving VLA policy.}
Given multi-view camera observations, a navigation command, and the
ego-vehicle state, we denote the resulting multimodal scene context by
$o$. A driving VLA policy maps $o$ to a continuous future trajectory
$\tau=\{w_t\}_{t=1}^{H}$, where $w_t$ represents the planned ego motion
at future step $t$. In our planner-based architecture, the VLM encodes
the multimodal context into latent features, while a conditional
diffusion planner models the trajectory distribution
$\pi_\theta(\tau\mid o)$. Following the common VLA training pipeline,
the policy is first initialized through supervised imitation learning
and subsequently improved through reinforcement learning.

\paragraph{Imitation learning.}
The trajectory policy is first trained on logged driving demonstrations.
Given a supervision target $\tau^\star$, we sample a diffusion step $t$
and Gaussian noise $\epsilon$, and construct the noisy trajectory
$x_t=\sqrt{\bar{\alpha}_t}\tau^\star+
\sqrt{1-\bar{\alpha}_t}\epsilon$.
The supervised fine-tuning objective is
\begin{equation}
\mathcal{L}_{\mathrm{SFT}}(\theta)
=
\mathbb{E}_{(o,\tau^\star),t,\epsilon}
\left[
\left\|
\epsilon-
\epsilon_\theta(x_t,t,o)
\right\|_2^2
\right].
\label{eq:sft_objective}
\end{equation}
Optimizing Eq.~\eqref{eq:sft_objective} on the logged trajectories
produces the GT-only imitation policy $\pi_{\mathrm{GT}}$. After convergence,
$\pi_{\mathrm{GT}}$ is frozen and used to generate rollout samples for candidate
selection. The same denoising objective is later used to learn selected
candidate trajectories and distilled improvement targets.

\paragraph{Group-relative policy optimization.}
Starting from an imitation policy, GRPO samples a group of $G$
trajectories
$\{\tau_i\}_{i=1}^{G}$ for the same scene and evaluates each trajectory
with an aggregate planning reward $R_i$. The trajectory-level policy
objective is written compactly as
\begin{equation}
\mathcal{L}_{\mathrm{GRPO}}(\theta)
=
-
\mathbb{E}
\left[
\frac{1}{G}
\sum_{i=1}^{G}
\widehat{A}_i
\log\pi_\theta(\tau_i\mid o)
\right]
+
\beta\mathcal{L}_{\mathrm{reg}},
\label{eq:grpo_objective}
\end{equation}
where $\log\pi_\theta(\tau_i\mid o)$ denotes the accumulated
log-probability of the denoising transitions that generate $\tau_i$,
and $\mathcal{L}_{\mathrm{reg}}$ denotes the behavior-cloning or KL
regularization used to limit policy drift.

Standard GRPO avoids an additional value network by estimating the
advantage from the rewards within the sampled group:
\begin{equation}
\widehat{A}_i
=
\frac{R_i-\mu_R}
{\sigma_R+\varepsilon},
\label{eq:grpo_advantage}
\end{equation}
where $\mu_R$ and $\sigma_R$ are the mean and standard deviation of
$\{R_i\}_{i=1}^{G}$.
Because Eq.~\eqref{eq:grpo_advantage} depends only on the aggregate
reward, it does not distinguish safety violations from meaningful
trade-offs among feasible trajectories. Section~\ref{sec:method}
therefore retains the same imitation and policy-optimization pipeline,
but redesigns trajectory selection and rollout credit assignment to
align multi-trajectory supervision with downstream optimization.

\section{Method}
\label{sec:method}

Our framework contains three stages, as shown in
Fig.~\ref{fig:framework}. We first train $\pi_{\mathrm{GT}}$ using logged
trajectories and freeze it for candidate selection. Policy-compatible multi-trajectory supervision (PC-MTS) then uses
high-quality candidates compatible with this policy to fine-tune a
trainable copy, producing the initialization $\pi_{\mathrm{initial}}$ for
GRPO. Feasibility-first Pareto GRPO (FF-PGRPO) next improves this policy through
feasibility-aware credit assignment. Finally, adaptive Pareto-guided policy
refinement (APR) repeatedly mines and distills improvements relative to the
latest policy. The evaluator and candidate pools are used only during training.

For clarity, we describe the metric hierarchy using NAVSIM. In v1, NC
and DAC determine hard feasibility, DDC is used as an additional
compliance guard, and EP, TTC, and comfort are compared within the
feasible region. NAVSIM v2 follows the same principle, adding DDC and
TLC to the feasibility and compliance checks and comparing its
continuous planning metrics only after these checks are satisfied.

\subsection{Policy-Compatible Multi-Trajectory Supervision}
\label{sec:policy_aware_selection}

We first filter the candidate pool by planning quality. Candidates must
satisfy the basic safety and compliance requirements, achieve a
sufficient aggregate score, and lie on the scene-wise Pareto front of
the component metrics. For NAVSIM v1, the Pareto comparison uses EP,
TTC, and comfort, while NC and DAC must remain valid and DDC is
protected. This retains high-quality alternatives with complementary
strengths, instead of allowing one strong component to conceal
degradation in another.

Planning quality alone does not determine whether a candidate is a
suitable target for $\pi_{\mathrm{GT}}$. We therefore sample multiple
rollouts from the frozen policy and compare each candidate with this
rollout bank. Compatibility is measured by the average waypoint
distance to the $K$ nearest policy rollouts. The acceptance threshold is
calibrated from held-out rollouts of the same policy and estimated
separately for each navigation command. This provides a sample-based
compatibility test without estimating the exact diffusion likelihood.

We additionally reject candidates with insufficient local feasibility.
For each candidate, we generate a small number of temporally coherent
nearby trajectories using the normal variation observed in
$\pi_{\mathrm{GT}}$ rollouts and require most of them to remain feasible.
The logged trajectory is always retained. During fine-tuning, each scene
contributes one target sampled from its logged trajectory and accepted
candidates, regardless of the size of its candidate pool. Fine-tuning a
copy of $\pi_{\mathrm{GT}}$ with the standard diffusion objective yields
$\pi_{\mathrm{initial}}$. The aim is not to maximize the imitation
checkpoint alone, but to provide a policy from which GRPO can reliably
sample useful trajectories.

\subsection{Feasibility-First Pareto GRPO}
\label{sec:pareto_grpo}

Starting from $\pi_{\mathrm{initial}}$, GRPO samples $G$ trajectories for
each scene and evaluates their aggregate and component scores. We first
construct a coherent scene reference $\tau^{\mathrm{ref}}$ from the
logged trajectory, the retained candidates, and the single-trajectory
prediction of $\pi_{\mathrm{initial}}$. The highest-scoring feasible
trajectory is selected, so all reference metrics come from the same
executable plan rather than from an unattainable combination of
different trajectories.

Credit assignment follows a feasibility-first hierarchy. In NAVSIM v1,
a rollout must satisfy NC and DAC. DDC must meet an absolute or
reference-relative guard, and a reference-relative EP floor prevents
the policy from increasing TTC simply by slowing down. Among the
rollouts that pass these checks, we construct the Pareto front over EP,
TTC, and comfort. We denote this admissible Pareto front by
$\mathcal P$.

The group quality signal is based on PDMS. For groups containing both
feasible and infeasible rollouts, we additionally penalize insufficient
EP and joint regression in EP and TTC relative to the scene reference.
When all rollouts are feasible, only the continuous component of PDMS
is normalized because the binary feasibility terms are already
satisfied. We continue to use $\widehat A_i$ for the resulting
group-normalized quality advantage.

When the group contains at least one feasible rollout, FF-PGRPO assigns
the advantage as
\begin{equation}
A_i
=
\begin{cases}
\max(\widehat A_i,0),
& \tau_i\in\mathcal P,\\
\min(\widehat A_i,0),
& \tau_i\text{ is feasible but }\tau_i\notin\mathcal P,\\
-\lambda_{\mathrm{unsafe}}+\min(\widehat A_i,0),
& \tau_i\text{ is infeasible}.
\end{cases}
\label{eq:hierarchical_advantage}
\end{equation}
Feasible rollouts that fail the reference guards are therefore prevented
from receiving positive credit, while unsafe rollouts receive a
stronger penalty. Aggregate quality controls the strength of an update;
feasibility, reference-level performance, and Pareto comparison control
whether a positive update is allowed. This prevents progress or
aggregate-score gains from compensating for safety regression.

\begin{figure*}[t]
    \centering

    \begin{subfigure}[t]{0.485\textwidth}
        \centering
        \includegraphics[width=\linewidth]{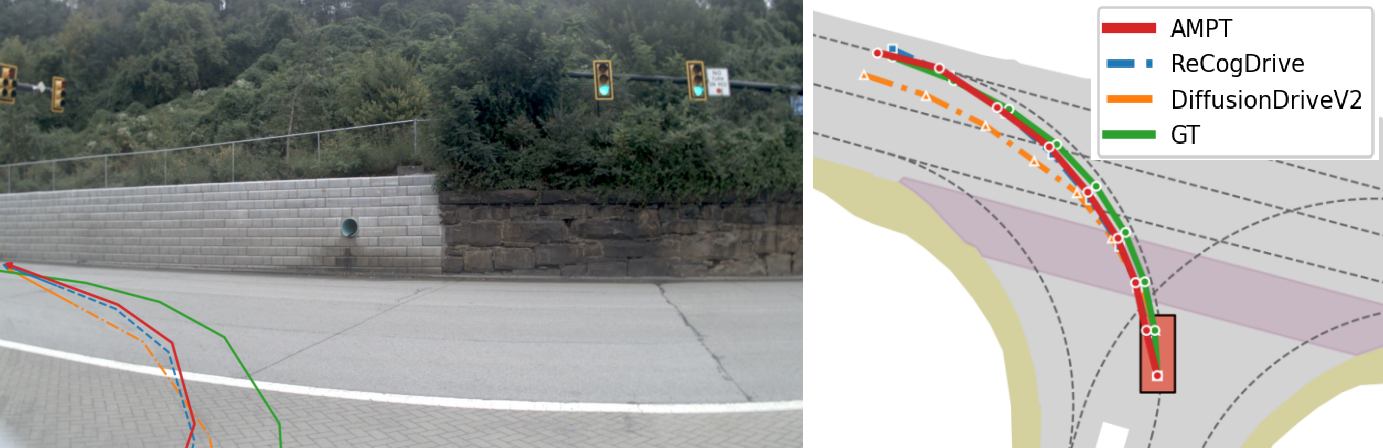}
        \caption{Driving command: Turn left.}
        \label{fig:subfig_a}
    \end{subfigure}
    \begin{subfigure}[t]{0.485\textwidth}
        \centering
        \includegraphics[width=\linewidth]{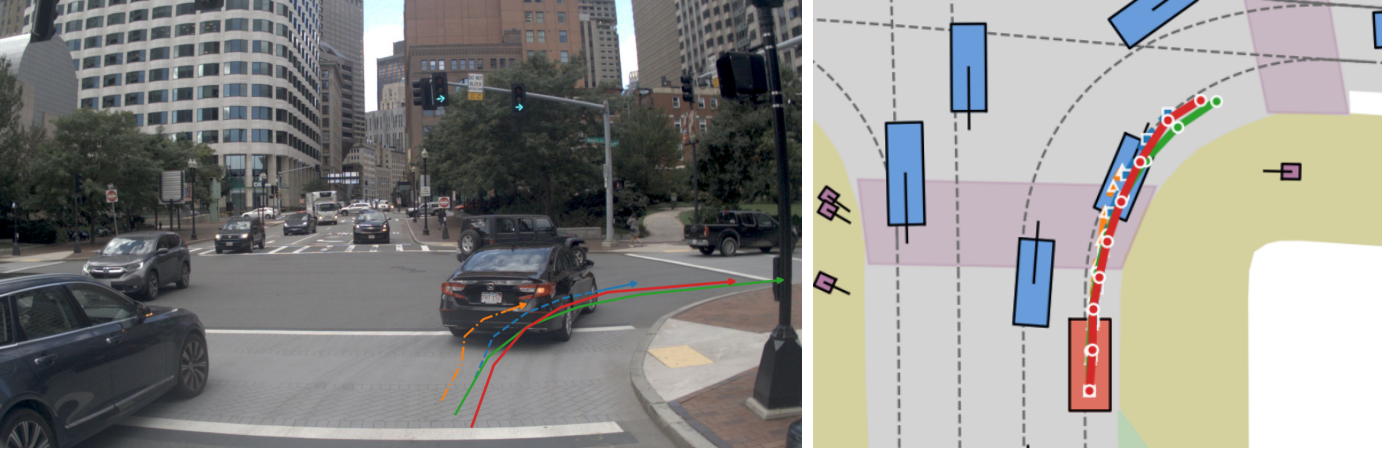}
        \caption{Driving command: Turn right.}
        \label{fig:subfig_b}
    \end{subfigure}

    \caption{Qualitative comparison of trajectories by different models in front-facing camera and bird’s eye view on different driving scenarios.}
    \label{fig:two_horizontal_subfigures}
\end{figure*}

If the whole group is infeasible, no rollout is treated as a positive
example. We assign non-positive credit according to violation severity
and add weak diffusion supervision from the safe scene reference, when
available. These groups are downweighted so repeated failures do not
dominate training. As in the baseline, the policy update is regularized
toward $\pi_{\mathrm{initial}}$ with behavior-cloning or KL terms. The
resulting policy first establishes or recovers feasibility and then
improves efficiency within the feasible region.

\subsection{Adaptive Pareto-Guided Policy Refinement}
\label{sec:iterative_refinement}

A single GRPO run may not absorb useful behaviors available from other
policy seeds, structured trajectory expanders, or policies specialized
for safety, progress, and trajectory regularity. We therefore build a
multi-source trajectory pool and reassess it against the current policy
at every refinement round.

For each scene, the current policy's single-trajectory prediction is
used as the baseline. A candidate can become a teacher only if it is
feasible, improves the aggregate score by a margin, and does not degrade
the protected component metrics. Among valid teachers, we prefer larger
improvements with smaller behavioral distance and lower curvature or
jerk, avoiding aggressive targets whose score gain comes with an
unnecessarily large policy shift.

The selected teacher is then interpolated with the current prediction.
We test a short descending list of interpolation ratios and re-evaluate
the resulting trajectories. The largest local target that remains
feasible, improves the aggregate score, preserves all protected metrics,
and stays compatible with current-policy rollouts is retained. This
step converts a potentially distant teacher into a verified local
improvement.

Accepted targets are weighted by their verified score gains and learned
with the same diffusion objective. Training also includes policy
retention on the current predictions and a distillation or KL term that
limits excessive drift. Scenes without a valid teacher contribute only
to retention. When safety-, progress-, and structure-oriented branches
are trained separately, we consolidate their salient, anchor-aligned
updates with a constrained task-vector merge; detailed merging rules are
left to the supplementary material.

After validation, the refined checkpoint becomes the current policy for
the next round. Its predictions and rollout bank are regenerated, and
the teacher set is rebuilt. Outdated teachers are removed, while
previously unsuitable candidates may enter after the policy becomes
closer to them. Neither the evaluator, the candidate pool, nor the
specialist branches are used at inference.

\section{Experiments}
\label{sec:experiments}

\subsection{Experimental Setup}
\label{sec:experimental_setup}

\paragraph{Benchmarks and evaluation protocol.}
We evaluate on NAVSIM v1 and NAVSIM v2 ~\cite{dauner2024navsim,cao2025pseudo}. NAVSIM v1 reports the Predictive Driver Model Score (PDMS), which combines no-at-fault collision (NC), drivable-area compliance (DAC), time to collision (TTC), comfort (C), and ego progress (EP). NAVSIM v2 extends the evaluation to EPDMS by including driving-direction compliance (DDC), traffic-light compliance (TLC), lane keeping (LK), history comfort (HC), and extended comfort (EC). We report the aggregate scores together with all component metrics, since a scalar improvement alone cannot reveal whether progress is obtained at the expense of safety or compliance. We also selected 658 challenging scenarios (where collisions or exceeding the drivable area occurred in all 5 rounds of sampling, resulting in a PDMS of 0) based on the ReCogDrive~\cite{li2025recogdrive} IL policy, to quantitatively analyze the safety improvements of our method.

\paragraph{Implementation details.}
To ensure fair comparisons and highlight our method's contribution to policy optimization, we reused the VLA backbone network InternVL3-2B~\cite{zhu2025internvl3} and diffusion planner from ReCogDrive. All controlled comparisons used the same initialization, trajectory representation, optimizer, training budget, and evaluation protocol. The GT-only policy and the multi-trajectory imitation variants are trained for 200 epochs. GRPO used 16 rollouts per scene for 10 epochs of optimization, and our method ultimately underwent 3 rounds of iterative optimization. All experiments were performed on 8 NVIDIA A800 GPUs; complete hyperparameters are provided in the supplementary materials.

\paragraph{Compared variants.}
We denote policy-compatible multi-trajectory supervision, feasibility-first
Pareto GRPO, and adaptive Pareto-guided policy refinement by \textbf{PC-MTS},
\textbf{FF-PGRPO}, and \textbf{APR}, respectively. To isolate trajectory-set
construction, \textbf{Score} retains candidates by aggregate score only,
while \textbf{Pareto} additionally applies component-wise Pareto filtering.
PC-MTS further removes candidates that are incompatible with the frozen
GT-only policy. In the Score/Pareto/PC-MTS comparison, every
imitation checkpoint is followed by the same original scalar GRPO.

\subsection{Main Results}
\label{sec:main_results}
\begin{table*}[t]
\centering
\caption{Single-trajectory planning results on NAVSIM v1.
$\uparrow$ indicates that higher is better.
The best and second-best results among the listed methods are shown
in bold and underlined, respectively.
Only the baselines selected for comparison are included.}
\label{tab:navsim_v1_main}
\small
\setlength{\tabcolsep}{4.0pt}
\begin{tabular*}{\textwidth}{
    @{\extracolsep{\fill}}lcccccc@{}
}
\toprule
Method
& NC$\uparrow$
& DAC$\uparrow$
& TTC$\uparrow$
& C$\uparrow$
& EP$\uparrow$
& \textbf{PDMS}$\uparrow$ \\
\midrule

LAW~\cite{li2025law}
& 96.4
& 95.4
& 88.7
& \secondscore{99.9}
& 81.7
& 84.6 \\

FSDrive~\cite{zeng2026futuresightdrive}
& 98.2
& 93.8
& 93.3
& \secondscore{99.9}
& 80.1
& 85.1 \\

Hydra-MDP~\cite{li2024hydra}
& 98.3
& 96.0
& 94.6
& \bestscore{100}
& 78.7
& 86.5 \\

DiffusionDrive~\cite{liao2025diffusiondrive}
& 98.2
& 96.2
& 94.7
& \bestscore{100}
& 82.2
& 88.1 \\

PWM~\cite{zhao2026forecasting}
& \secondscore{98.6}
& 95.9
& 95.4
& \bestscore{100}
& 81.8
& 88.1 \\

AutoVLA~\cite{zhou2025autovla}
& 98.4
& 95.6
& \bestscore{98.0}
& \secondscore{99.9}
& 81.9
& 89.1 \\

DriveVLA-W0~\cite{li2025drivevla}
& \bestscore{98.7}
& \bestscore{99.1}
& 95.3
& 99.3
& 83.3
& 90.2 \\

GoalFlow~\cite{xing2025goalflow}
& 98.4
& \secondscore{98.3}
& 94.6
& \bestscore{100}
& 85.0
& 90.3 \\

Curious-VLA~\cite{chen2026curiousvla}
& 98.4
& 96.9
& \secondscore{97.9}
& 98.1
& \bestscore{88.5}
& 90.3 \\

AdathinkDrive~\cite{luo2025adathinkdrive}
& 98.4
& 97.8
& 95.2
& \bestscore{100}
& 84.4
& 90.3 \\

ReCogDrive~\cite{li2025recogdrive}
& 97.9
& 97.3
& 94.9
& \bestscore{100}
& 87.3
& 90.8 \\

DiffusionDriveV2~\cite{zou2025diffusiondrivev2}
& 98.3
& 97.9
& 94.8
& \secondscore{99.9}
& \secondscore{87.5}
& \secondscore{91.2} \\

\midrule

\textbf{AMPT (Ours)}
& 98.5
& 98.0
& 96.0
& \bestscore{100}
& 86.7
& \bestscore{91.4} \\

\bottomrule
\end{tabular*}
\end{table*}

\begin{table}[t]
\centering
\caption{Effect of trajectory-set construction. Every imitation
checkpoint is followed by the same original scalar GRPO.}
\label{tab:data_construction}
\footnotesize
\setlength{\tabcolsep}{5pt}
\begin{tabular}{lccc}
\toprule
Data & IL & GRPO & $\Delta$ \\
\midrule
GT & 86.4 & 90.3 & $+3.9$ \\
Score & 87.2 & 85.8 & $-1.4$ \\
Pareto & \textbf{87.5} & 86.7 & $-0.8$ \\
PC-MTS & 86.9 & \textbf{91.1} & \textbf{$+4.2$} \\
\bottomrule
\end{tabular}
\end{table}

\begin{table*}[t]
\centering
\caption{Single-trajectory planning results on NAVSIM v2.
$\uparrow$ indicates that higher is better.
NC, DAC, DDC, and TLC are multiplicative compliance metrics,
while TTC, EP, LK, HC, and EC form the weighted component of EPDMS.
The best and second-best results among the listed methods are shown
in bold and underlined, respectively.}
\label{tab:navsim_v2_main}
\small
\setlength{\tabcolsep}{3.5pt}
\begin{tabular*}{\textwidth}{
    @{\extracolsep{\fill}}lcccccccccc@{}
}
\toprule
Method
& NC$\uparrow$
& DAC$\uparrow$
& DDC$\uparrow$
& TLC$\uparrow$
& TTC$\uparrow$
& EP$\uparrow$
& LK$\uparrow$
& HC$\uparrow$
& EC$\uparrow$
& \textbf{EPDMS}$\uparrow$ \\
\midrule

TransFuser~\cite{chitta2022transfuser}
& 97.7
& 92.8
& 98.3
& \secondscore{99.9}
& 92.8
& 79.2
& 67.6
& \bestscore{100}
& \secondscore{95.3}
& 77.8 \\

ReCogDrive~\cite{li2025recogdrive}
& 98.3
& 95.2
& \bestscore{99.5}
& 99.8
& 97.5
& 87.1
& 96.6
& \secondscore{98.3}
& 86.5
& 83.6 \\

Hydra-MDP++~\cite{li2025hydra}
& \bestscore{98.8}
& 97.8
& 99.1
& \bestscore{100}
& 95.3
& 84.0
& 70.1
& \bestscore{100}
& \bestscore{96.8}
& 84.1 \\

DiffusionDrive~\cite{liao2025diffusiondrive}
& 98.2
& 95.9
& \secondscore{99.4}
& 99.8
& 97.3
& 87.5
& 96.8
& \secondscore{98.3}
& 87.7
& 84.5 \\

Curious-VLA~\cite{chen2026curiousvla}
& 98.4
& 96.9
& 99.2
& 99.8
& \secondscore{97.9}
& 88.5
& \secondscore{96.9}
& 98.1
& 81.5
& 85.3 \\

DiffusionDriveV2~\cite{zou2025diffusiondrivev2}
& 97.7
& 96.6
& 99.2
& 99.8
& 97.2
& \secondscore{88.9}
& 96.0
& 97.8
& 91.0
& 85.5 \\

DriveVLA-W0~\cite{li2025drivevla}
& \secondscore{98.5}
& \bestscore{99.1}
& 98.0
& 99.7
& \bestscore{98.1}
& 86.4
& 93.2
& 97.9
& 58.9
& 86.1 \\

Drive-JEPA~\cite{wang2026drivejepa}
& 98.4
& \secondscore{98.6}
& 99.1
& 99.8
& 97.8
& 88.4
& \bestscore{97.6}
& 97.9
& 84.8
& \secondscore{87.8} \\

\midrule

\textbf{AMPT (Ours)}
& 98.4
& 97.7
& 99.0
& 99.7
& 97.7
& \bestscore{89.4}
& 91.7
& 97.1
& 87.7
& \bestscore{89.1} \\

\bottomrule
\end{tabular*}
\end{table*}

\paragraph{NAVSIM v1.}
As shown in Table ~\ref{tab:navsim_v1_main}, AMPT achieves a PDMS score of 91.4, reaching state-of-the-art (SOTA) among all compared methods, 0.2 points higher than DiffusionDriveV2. The margin is modest, but the component profile is informative, and our method does not rely on an additional trained selector module to choose the optimal trajectory from multiple candidate trajectory outputs. Relative to DiffusionDriveV2, our policy improves NC, DAC, and TTC, including a 1.2-point gain in TTC, while retaining competitive EP. The result is therefore not obtained by pushing the policy toward a more aggressive progress profile. Instead, it reflects a stronger balance between feasibility, safety margin, and driving efficiency, which is consistent with our proposed feasibility-first Pareto optimization philosophy.

\paragraph{NAVSIM v2.}
Table~\ref{tab:navsim_v2_main} evaluates the same policy under the richer EPDMS protocol. AMPT reaches 89.1 EPDMS, improving over the strongest compared baseline, Drive-JEPA, by 1.3 points. The gain is primarily associated with higher EP and EC, while NC, DDC, TLC, and TTC remain close to their saturated ranges. Although the method does not maximize every individual component, it obtains the strongest aggregate result under the expanded set of safety, compliance, progress, and comfort requirements. This confirms that the proposed approach is not specific to the metric composition of NAVSIM v1.

\subsection{Does Better Imitation Performance Lead to Better GRPO Initialization?}
\label{sec:trajectory_set_ablation}

We examine whether a higher-scoring imitation policy also provides a better initialization for GRPO. We compare GT supervision with Score, Pareto, and PC-MTS. All variants use the same imitation objective and are subsequently optimized with the same scalar GRPO, isolating the effect of trajectory-set construction.

Table~\ref{tab:data_construction} shows a clear reversal between imitation and post-training. Score and Pareto improve the imitation checkpoint to 87.2 and 87.5, respectively, but degrade to 85.8 and 86.7 after GRPO. In contrast, PC-MTS reaches only 86.9 after imitation yet improves to 91.1 after GRPO. Thus, the trajectory set that produces the best imitation checkpoint does not yield the best optimized policy. This result directly supports the supervision--optimization mismatch. Aggregate score and component-wise Pareto quality identify desirable plans, but not whether learning them preserves a feasible and improvable rollout distribution. Trajectory supervision should therefore be evaluated by its effect on downstream policy optimization rather than by imitation performance alone.

\subsection{Ablation and Further Analysis}
\label{sec:ablation_analysis}

\paragraph{Contribution of the three stages.}
Table~\ref{tab:stage_ablation} separates their roles. PC-MTS gives a
modest direct gain, consistent with its role as an initialization
mechanism. FF-PGRPO contributes most: relative to the GT-only policy, it
raises DAC from 94.7 to 98.0, TTC from 94.2 to 95.8, and EP from 80.9 to
85.9, reaching 91.0 PDMS. The joint gains in safety-sensitive components
and progress show that efficiency is not obtained by relaxing feasibility.

Adding PC-MTS reaches 91.1 PDMS. Together with
Table~\ref{tab:data_construction}, this shows that its main value is to
prevent high-score expansion from producing a poor GRPO initialization.
APR further raises PDMS to 91.45 by increasing EP from 86.0 to 86.7 while
NC, DAC, TTC, and comfort remain unchanged, recovering residual efficiency
without undoing the safety gains of GRPO.

\begin{table}[t]
\centering
\caption{\textbf{Ablation of PC-MTS, FF-PGRPO, and APR on NAVSIM v1.}
Higher is better; the best and second-best values are bold and underlined.}
\label{tab:stage_ablation}
\small
\setlength{\tabcolsep}{3.0pt}
\resizebox{\columnwidth}{!}{%
\begin{tabular}{ccccccccc}
\toprule
\multicolumn{3}{c}{Training Stage}
& \multicolumn{6}{c}{NAVSIM v1} \\
\cmidrule(lr){1-3}
\cmidrule(lr){4-9}
PC-MTS
& FF-PGRPO
& APR
& NC$\uparrow$
& DAC$\uparrow$
& TTC$\uparrow$
& C$\uparrow$
& EP$\uparrow$
& PDMS$\uparrow$ \\
\midrule

--
& --
& --
& 98.1
& 94.7
& 94.2
& 100
& 80.9
& 86.5 \\

\checkmark
& --
& --
& 98.2
& \secondscore{95.2}
& 94.5
& 100
& 81.3
& 86.9 \\

--
& \checkmark
& --
& \secondscore{98.4}
& \bestscore{98.0}
& \secondscore{95.8}
& 100
& 85.93
& 91.0 \\

\checkmark
& \checkmark
& --
& \bestscore{98.5}
& \bestscore{98.0}
& \bestscore{96.0}
& 100
& \secondscore{86.0}
& \secondscore{91.1} \\

\midrule

\checkmark
& \checkmark
& \checkmark
& \bestscore{98.5}
& \bestscore{98.0}
& \bestscore{96.0}
& 100
& \bestscore{86.7}
& \bestscore{91.4} \\

\bottomrule
\end{tabular}%
}
\end{table}

\paragraph{Adaptive policy refinement.}
Starting from the PC-MTS+FF-PGRPO policy, three refinement rounds improve PDMS from 91.10 to 91.21, 91.37, and 91.45, respectively (Table~\ref{tab:iterative_results}). The increments are small, as expected for refinement of an already strong policy, but remain monotonic across the three rounds. This progression supports the use of dynamic teacher reconstruction: after each update, obsolete teachers areremoved and the remaining candidate pool is searched for improvementsrelative to the new policy, rather than repeatedly fitting a static set.

\begin{table}[t]
\centering
\caption{Performance across iterative refinement rounds. Round 0 is the
PC-MTS+FF-PGRPO policy before APR.}
\label{tab:iterative_results}
\small
\setlength{\tabcolsep}{6pt}
\begin{tabular}{lcc}
\toprule
Policy & PDMS & Gain per round \\
\midrule
Round 0 (before APR) & 91.10 & -- \\
Round 1         & 91.21 & $+0.11$ \\
Round 2         & 91.37 & $+0.16$ \\
Round 3         & \textbf{91.45} & $+0.08$ \\
\bottomrule
\end{tabular}
\end{table}

\paragraph{Failure recovery.}
The aggregate score improvements also translate to a broader correction of difficult scenes. On the fixed set of 658 initial failures, the original scalar GRPO recovers 367 scenes, whereas AMPT recovers 440 . This increases the recovery rate from 55.8\% to 66.9\%, corresponding to 73 additional recovered scenes and an absolute gain of 11.1 percentage points. The improvement is therefore not limited to marginal score changes on already successful cases; it also reflects a larger ability to move failed scenes back into the feasible region.

Taken together, the experiments identify distinct but complementary roles for the three stages. PC-MTS prevents high-scoring trajectory expansion from creating a poor GRPO initialization; FF-PGRPO provides the main online gain by improving progress inside the feasible region; and APR continues to extract policy-relative improvements after GRPO has converged. The failure-recovery analysis further shows that these gains extend to the hard tail of the evaluation set.

\section{Conclusion}
\label{sec:conclusion}

We studied how multi-trajectory supervision can be made useful for subsequent policy optimization in VLA driving. Our results expose a supervision--optimization mismatch: trajectory sets that improve the imitation checkpoint can still produce a substantially worse GRPO initialization. We address this mismatch with AMPT, which combines PC-MTS, FF-PGRPO, and APR. The resulting single-trajectory policy achieves 91.4 PDMS on NAVSIM v1 and 89.1 EPDMS on NAVSIM v2, while recovering 440 of 658 initially failed scenes. These results indicate that expanded supervision should be judged by the policy distribution it induces for downstream optimization, rather than by trajectory scores in isolation. The current compatibility test relies on finite policy samples, and teacher validation depends on an offline planning evaluator. Future work will study learned compatibility estimates and extend the framework to interactive closed-loop environments.

\bibliography{aaai2027}

\clearpage
\begingroup
\def\AMPTEmbeddedSupplement{}
\setlength{\textfloatsep}{10pt plus 2pt minus 2pt}
\setlength{\floatsep}{10pt plus 2pt minus 2pt}
\setlength{\intextsep}{10pt plus 2pt minus 2pt}
\setcounter{dbltopnumber}{2}
\renewcommand{\dbltopfraction}{0.95}
\renewcommand{\dblfloatpagefraction}{0.30}
\makeatletter
\setlength{\@dblfptop}{0pt}
\setlength{\@dblfpsep}{10pt plus 2pt minus 2pt}
\setlength{\@dblfpbot}{0pt plus 1fil}
\makeatother
\ifdefined\AMPTEmbeddedSupplement
\else
\documentclass[letterpaper]{article} 
\usepackage[submission]{aaai2027} 
\usepackage[hyphens]{url} 
\usepackage{graphicx} 
\urlstyle{rm} 
\def\UrlFont{\rm} 
\usepackage{natbib} 
\usepackage{caption} 
\usepackage{amsmath}
\usepackage{amssymb}
\usepackage{booktabs}
\usepackage{algorithm}
\usepackage{algorithmic}
\frenchspacing 

\pdfinfo{
/TemplateVersion (2027.1)
}

\title{Supplementary Material for ``Aligning Multi-Trajectory Supervision
with Policy Optimization for VLA Driving''}

\author{
    Anonymous Submission
}
\affiliations{
}

\begin{document}

\maketitle
\fi

\providecommand{\AMPT}{\textsc{AMPT}}
\providecommand{\PCMTS}{\textsc{PC-MTS}}
\providecommand{\FFPGRPO}{\textsc{FF-PGRPO}}
\providecommand{\APR}{\textsc{APR}}
\providecommand{\NA}{--}
\graphicspath{{figures/generated/}}
\setcounter{secnumdepth}{1}

\ifdefined\AMPTEmbeddedSupplement
    \newcommand{\supplementclearpage}{}
    \newcommand{\supplementbibliography}{}
    \newcommand{\supplementend}{}
    \newcommand{\supplementnotationsize}{\small}
    \newcommand{\supplementalgorithmsize}{\footnotesize}
    \newcommand{\supplementonecolumn}{}
    \newcommand{\supplementtwocolumn}{}
    \newcommand{\supplementfloatbarrier}{\FloatBarrier}
    \newcommand{\supplementalgorithmgroupbegin}{\begin{figure*}[!tp]}
    \newcommand{\supplementalgorithmgroupend}{\end{figure*}}
    \newenvironment{supplementalgorithm}
        {\begin{minipage}{\textwidth}\captionsetup{type=algorithm}}
        {\end{minipage}\par\vspace{6pt}}
    \newcommand{\supplementalgorithmthree}[1]{%
        \twocolumn[{\begin{minipage}{\textwidth}
        \captionsetup{type=algorithm}#1
        \end{minipage}}]}
    \newcommand{\supplementembeddedtitle}{%
        \begin{center}
        {\Large\bfseries Supplementary Material for ``Aligning Multi-Trajectory
        Supervision with Policy Optimization for VLA Driving''\par}
        \end{center}}
\else
    \newcommand{\supplementclearpage}{\clearpage}
    \newcommand{\supplementbibliography}{\bibliography{aaai2027}}

@inproceedings{tian2024drivevlm,
  title={{DriveVLM}: The Convergence of Autonomous Driving and Large Vision-Language Models},
  author={Tian, Xiaoyu and Gu, Junru and Li, Bailin and Liu, Yicheng and Wang, Yang and Zhao, Zhiyong and Zhan, Kun and Jia, Peng and Lang, XianPeng and Zhao, Hang},
  booktitle={Proceedings of the 8th Conference on Robot Learning},
  series={Proceedings of Machine Learning Research},
  volume={270},
  pages={4698--4726},
  publisher={PMLR},
  year={2025},
  url={https://proceedings.mlr.press/v270/tian25c.html}
}

@inproceedings{wang2025omnidrive,
  title={{OmniDrive}: A Holistic Vision-Language Dataset for Autonomous Driving with Counterfactual Reasoning},
  author={Wang, Shihao and Yu, Zhiding and Jiang, Xiaohui and Lan, Shiyi and Shi, Min and Chang, Nadine and Kautz, Jan and Li, Ying and Alvarez, Jose M.},
  booktitle={Proceedings of the IEEE/CVF Conference on Computer Vision and Pattern Recognition},
  pages={22442--22452},
  year={2025}
}

@article{hwang2024emma,
  title={{EMMA}: End-to-End Multimodal Model for Autonomous Driving},
  author={Hwang, Jyh-Jing and Xu, Runsheng and Lin, Hubert and Hung, Wei-Chih and Ji, Jingwei and Choi, Kristy and Huang, Di and He, Tong and Covington, Paul and Sapp, Benjamin and Zhou, Yin and Guo, James and Anguelov, Dragomir and Tan, Mingxing},
  journal={arXiv preprint arXiv:2410.23262},
  year={2024}
}

@article{zhou2025opendrivevla,
  title={{OpenDriveVLA}: Towards End-to-End Autonomous Driving with Large Vision Language Action Model},
  author={Zhou, Xingcheng and Han, Xuyuan and Yang, Feng and Ma, Yunpu and Knoll, Alois C.},
  journal={arXiv preprint arXiv:2503.23463},
  year={2025}
}

@inproceedings{chi2023diffusionpolicy,
  title={Diffusion Policy: Visuomotor Policy Learning via Action Diffusion},
  author={Chi, Cheng and Xu, Zhenjia and Feng, Siyuan and Cousineau, Eric and Du, Yilun and Burchfiel, Benjamin and Tedrake, Russ and Song, Shuran},
  booktitle={Robotics: Science and Systems},
  year={2023},
  doi={10.15607/RSS.2023.XIX.026}
}

@inproceedings{liao2025diffusiondrive,
  title={{DiffusionDrive}: Truncated Diffusion Model for End-to-End Autonomous Driving},
  author={Liao, Bencheng and Chen, Shaoyu and Yin, Haoran and Jiang, Bo and Wang, Cheng and Yan, Sixu and Zhang, Xinbang and Li, Xiangyu and Zhang, Ying and Zhang, Qian and Wang, Xinggang},
  booktitle={Proceedings of the IEEE/CVF Conference on Computer Vision and Pattern Recognition},
  pages={12037--12047},
  year={2025}
}

@article{wang2025hmad,
  title={{HMAD}: Advancing E2E Driving with Anchored Offset Proposals and Simulation-Supervised Multi-Target Scoring},
  author={Wang, Bin and Li, Pingjun and Liu, Jinkun and Cheng, Jun and Lei, Hailong and Rong, Yinze and Gao, Huan-ang and Chen, Kangliang and Pan, Xing and Gu, Weihao},
  journal={arXiv preprint arXiv:2505.23129},
  year={2025}
}

@article{wang2026drivejepa,
  title={{Drive-JEPA}: Video JEPA Meets Multimodal Trajectory Distillation for End-to-End Driving},
  author={Wang, Linhan and Yang, Zichong and Bai, Chen and Zhang, Guoxiang and Liu, Xiaotong and Zheng, Xiaoyin and Long, Xiao-Xiao and Lu, Chang-Tien and Lu, Cheng},
  journal={arXiv preprint arXiv:2601.22032},
  year={2026}
}

@article{chen2026curiousvla,
  title={Devil is in Narrow Policy: Unleashing Exploration in Driving {VLA} Models},
  author={Chen, Canyu and Yang, Yuguang and Tan, Zhewen and Wang, Yizhi and Zhan, Ruiyi and Liu, Haiyan and Mao, Xuanyao and Bao, Jason and Tang, Xinyue and Yang, Linlin and Sun, Bingchuan and Wang, Yan and Zhang, Baochang},
  journal={arXiv preprint arXiv:2603.06049},
  year={2026}
}

@article{li2025recogdrive,
  title={{ReCogDrive}: A Reinforced Cognitive Framework for End-to-End Autonomous Driving},
  author={Li, Yongkang and Xiong, Kaixin and Guo, Xiangyu and Li, Fang and Yan, Sixu and Xu, Gangwei and Zhou, Lijun and Chen, Long and Sun, Haiyang and Wang, Bing and Chen, Guang and Ye, Hangjun and Liu, Wenyu and Wang, Xinggang},
  journal={arXiv preprint arXiv:2506.08052},
  year={2025}
}

@article{jiang2025alphadrive,
  title={{AlphaDrive}: Unleashing the Power of VLMs in Autonomous Driving via Reinforcement Learning and Reasoning},
  author={Jiang, Bo and Chen, Shaoyu and Zhang, Qian and Liu, Wenyu and Wang, Xinggang},
  journal={arXiv preprint arXiv:2503.07608},
  year={2025}
}

@article{tang2025planr1,
  title={{Plan-R1}: Safe and Feasible Trajectory Planning as Language Modeling},
  author={Tang, Xiaolong and Kan, Meina and Shan, Shiguang and Chen, Xilin},
  journal={arXiv preprint arXiv:2505.17659},
  year={2025}
}

@article{zou2025diffusiondrivev2,
  title={{DiffusionDriveV2}: Reinforcement Learning-Constrained Truncated Diffusion Modeling in End-to-End Autonomous Driving},
  author={Zou, Jialv and Chen, Shaoyu and Liao, Bencheng and Zheng, Zhiyu and Song, Yuehao and Zhang, Lefei and Zhang, Qian and Liu, Wenyu and Wang, Xinggang},
  journal={arXiv preprint arXiv:2512.07745},
  year={2025}
}

@article{yao2026had,
  title={{HAD}: Combining Hierarchical Diffusion with Metric-Decoupled RL for End-to-End Driving},
  author={Yao, Wenhao and Sun, Xinglong and Li, Zhenxin and Lan, Shiyi and Wang, Zi and Alvarez, Jose M. and Wu, Zuxuan},
  journal={arXiv preprint arXiv:2604.03581},
  year={2026}
}

@article{jiao2025evadrive,
  title={{EvaDrive}: Evolutionary Adversarial Policy Optimization for End-to-End Autonomous Driving},
  author={Jiao, Siwen and Qian, Kangan and Ye, Hao and Zhong, Yang and Luo, Ziang and Jiang, Sicong and Huang, Zilin and Fang, Yangyi and Miao, Jinyu and Fu, Zheng and Wang, Yunlong and Jiang, Kun and Yang, Diange and Fan, Rui and Peng, Baoyun},
  journal={arXiv preprint arXiv:2508.09158},
  year={2025}
}

@article{ang2026clover,
  title={{CLOVER}: Closed-Loop Value Estimation \& Ranking for End-to-End Autonomous Driving Planning},
  author={Ang, Sining and Yang, Yuguang and Chen, Canyu and Wang, Yan},
  journal={arXiv preprint arXiv:2605.15120},
  year={2026}
}

@inproceedings{kumar2019bear,
  title={Stabilizing Off-Policy Q-Learning via Bootstrapping Error Reduction},
  author={Kumar, Aviral and Fu, Justin and Soh, Matthew and Tucker, George and Levine, Sergey},
  booktitle={Advances in Neural Information Processing Systems},
  volume={32},
  pages={11761--11771},
  year={2019}
}

@inproceedings{laroche2019spibb,
  title={Safe Policy Improvement with Baseline Bootstrapping},
  author={Laroche, Romain and Trichelair, Paul and Tachet des Combes, R{\'e}mi},
  booktitle={Proceedings of the 36th International Conference on Machine Learning},
  series={Proceedings of Machine Learning Research},
  volume={97},
  pages={3652--3661},
  publisher={PMLR},
  year={2019}
}

@article{nair2020awac,
  title={{AWAC}: Accelerating Online Reinforcement Learning with Offline Datasets},
  author={Nair, Ashvin and Gupta, Abhishek and Dalal, Murtaza and Levine, Sergey},
  journal={arXiv preprint arXiv:2006.09359},
  year={2020}
}

@inproceedings{dauner2024navsim,
  title={{NAVSIM}: Data-Driven Non-Reactive Autonomous Vehicle Simulation and Benchmarking},
  author={Dauner, Daniel and Hallgarten, Marcel and Li, Tianyu and Weng, Xinshuo and Huang, Zhiyu and Yang, Zetong and Li, Hongyang and Gilitschenski, Igor and Ivanovic, Boris and Pavone, Marco and Geiger, Andreas and Chitta, Kashyap},
  booktitle={Advances in Neural Information Processing Systems},
  volume={37},
  year={2024},
  note={Datasets and Benchmarks Track},
  doi={10.52202/079017-0902},
  url={https://proceedings.neurips.cc/paper_files/paper/2024/hash/32768f7faf1995026ef9821c696f3404-Abstract-Datasets_and_Benchmarks_Track.html}
}

@article{cao2025pseudo,
  title={Pseudo-simulation for autonomous driving},
  author={Cao, Wei and Hallgarten, Marcel and Li, Tianyu and Dauner, Daniel and Gu, Xunjiang and Wang, Caojun and Miron, Yakov and Aiello, Marco and Li, Hongyang and Gilitschenski, Igor and others},
  journal={arXiv preprint arXiv:2506.04218},
  year={2025}
}

@article{zhu2025internvl3,
  title={Internvl3: Exploring advanced training and test-time recipes for open-source multimodal models},
  author={Zhu, Jinguo and Wang, Weiyun and Chen, Zhe and Liu, Zhaoyang and Ye, Shenglong and Gu, Lixin and Tian, Hao and Duan, Yuchen and Su, Weijie and Shao, Jie and others},
  journal={arXiv preprint arXiv:2504.10479},
  year={2025}
}

@inproceedings{li2025law,
  title={Enhancing End-to-End Autonomous Driving with Latent World Model},
  author={Li, Yingyan and Fan, Lue and He, Jiawei and Wang, Yuqi and Chen, Yuntao and Zhang, Zhaoxiang and Tan, Tieniu},
  booktitle={International Conference on Learning Representations},
  year={2025},
  url={https://proceedings.iclr.cc/paper_files/paper/2025/hash/6aa4967920e495e90aeeaa3acf18d019-Abstract-Conference.html}
}

@inproceedings{zhou2025autovla,
  title={{AutoVLA}: A Vision-Language-Action Model for End-to-End Autonomous Driving with Adaptive Reasoning and Reinforcement Fine-Tuning},
  author={Zhou, Zewei and Cai, Tianhui and Zhao, Seth and Zhang, Yun and Huang, Zhiyu and Zhou, Bolei and Ma, Jiaqi},
  booktitle={Advances in Neural Information Processing Systems},
  volume={38},
  year={2025},
  url={https://proceedings.neurips.cc/paper_files/paper/2025/hash/2843fccca5bedd369a4764848b9bd546-Abstract-Conference.html}
}

@article{chitta2022transfuser,
  title={Transfuser: Imitation with transformer-based sensor fusion for autonomous driving},
  author={Chitta, Kashyap and Prakash, Aditya and Jaeger, Bernhard and Yu, Zehao and Renz, Katrin and Geiger, Andreas},
  journal={IEEE transactions on pattern analysis and machine intelligence},
  volume={45},
  number={11},
  pages={12878--12895},
  year={2022},
  publisher={IEEE}
}

@article{li2025drivevla,
  title={DriveVLA-W0: World models amplify data scaling law in autonomous driving},
  author={Li, Yingyan and Shang, Shuyao and Liu, Weisong and Zhan, Bing and Wang, Haochen and Wang, Yuqi and Chen, Yuntao and Wang, Xiaoman and An, Yasong and Tang, Chufeng and others},
  journal={arXiv preprint arXiv:2510.12796},
  year={2025}
}

@article{li2025hydra,
  title={Hydra-mdp++: Advancing end-to-end driving via expert-guided hydra-distillation},
  author={Li, Kailin and Li, Zhenxin and Lan, Shiyi and Xie, Yuan and Zhang, Zhizhong and Liu, Jiayi and Wu, Zuxuan and Yu, Zhiding and Alvarez, Jose M},
  journal={arXiv preprint arXiv:2503.12820},
  year={2025}
}

@article{li2024hydra,
  title={Hydra-mdp: End-to-end multimodal planning with multi-target hydra-distillation},
  author={Li, Zhenxin and Li, Kailin and Wang, Shihao and Lan, Shiyi and Yu, Zhiding and Ji, Yishen and Li, Zhiqi and Zhu, Ziyue and Kautz, Jan and Wu, Zuxuan and others},
  journal={arXiv preprint arXiv:2406.06978},
  year={2024}
}

@article{zeng2026futuresightdrive,
  title={Futuresightdrive: Thinking visually with spatio-temporal cot for autonomous driving},
  author={Zeng, Shuang and Chang, Xinyuan and Xie, Mengwei and Liu, Xinran and Bai, Yifan and Pan, Zheng and Xu, Mu and Wei, Xing},
  journal={Advances in Neural Information Processing Systems},
  volume={38},
  pages={67299--67318},
  year={2026}
}

@article{zhao2026forecasting,
  title={From forecasting to planning: Policy world model for collaborative state-action prediction},
  author={Zhao, Zhida and Fu, Talas and Wang, Yifan and Wang, Lijun and Lu, Huchuan},
  journal={Advances in Neural Information Processing Systems},
  volume={38},
  pages={134585--134611},
  year={2026}
}

@inproceedings{xing2025goalflow,
  title={Goalflow: Goal-driven flow matching for multimodal trajectories generation in end-to-end autonomous driving},
  author={Xing, Zebin and Zhang, Xingyu and Hu, Yang and Jiang, Bo and He, Tong and Zhang, Qian and Long, Xiaoxiao and Yin, Wei},
  booktitle={Proceedings of the Computer Vision and Pattern Recognition Conference},
  pages={1602--1611},
  year={2025}
}

@article{luo2025adathinkdrive,
  title={Adathinkdrive: Adaptive thinking via reinforcement learning for autonomous driving},
  author={Luo, Yuechen and Li, Fang and Xu, Shaoqing and Lai, Zhiyi and Yang, Lei and Chen, Qimao and Luo, Ziang and Xie, Zixun and Jiang, Shengyin and Liu, Jiaxin and others},
  journal={arXiv preprint arXiv:2509.13769},
  year={2025}
}
    \newcommand{\supplementend}{\end{document}}
    \newcommand{\supplementnotationsize}{}
    \newcommand{\supplementalgorithmsize}{\small}
    \newcommand{\supplementonecolumn}{\onecolumn}
    \newcommand{\supplementtwocolumn}{\twocolumn}
    \newcommand{\supplementfloatbarrier}{}
    \newcommand{\supplementalgorithmgroupbegin}{}
    \newcommand{\supplementalgorithmgroupend}{}
    \newenvironment{supplementalgorithm}{\begin{algorithm}[H]}{\end{algorithm}}
    \newcommand{\supplementalgorithmthree}[1]{%
        \begin{algorithm}[H]#1\end{algorithm}}
    \newcommand{\supplementembeddedtitle}{}
\fi

\newcommand{\supplementalgorithmheading}{%
\section{Algorithms}
\label{app:algorithms}
\supplementnotationsize
\paragraph{Notation.}
For scene $s$, $o_s$ is the multimodal input, $\tau_s^{\mathrm{gt}}$ is its
logged trajectory, and $\mathcal C_s$ is the candidate pool. The planning
evaluator returns an aggregate score $R(\tau)$, component metrics, and a
feasibility indicator $F(\tau)\in\{0,1\}$. $\operatorname{PF}(\cdot)$ denotes
the non-dominated subset under the active Pareto objectives.
$\pi_{\mathrm{GT}}$, $\pi_{\mathrm{initial}}$, and $\pi_{\mathrm{FF}}$ denote
the GT-only, PC-MTS, and FF-PGRPO policies, respectively. During APR,
$\pi^{(k)}$ is the policy at refinement round $k$.
$\operatorname{Guard}(\tau,\tau^{\mathrm{ref}})$ denotes the
reference-relative progress and compliance checks, and
$\operatorname{Audit}(\tau,\tau^{\mathrm{cur}})$ denotes feasibility,
positive score margin, and non-regression of protected metrics.}

\appendix
\supplementembeddedtitle

\noindent
\fbox{%
\parbox{0.96\columnwidth}{%
\textbf{\large Scope of the Study.}
\textbf{The proposed framework AMPT improves VLA driving from the perspectives of
training-data construction and policy optimization. It introduces no additional
external driving data, does not separately fine-tune the VLM backbone, and does
not increase the deployed policy's parameter count. The study focuses on how
multi-trajectory candidates can be selected and exploited to improve trajectory-policy optimization, rather than on feature extraction, representation learning, feature fusion, or multimodal semantic alignment. These research directions are orthogonal to and fully compatible with the proposed framework.}%
}}
\vspace{6pt}

\section{Additional Method Details}
\label{app:method-details}

This supplement provides the complete implementation details and extended
analysis of the three stages in \AMPT: policy-compatible multi-trajectory
supervision (\PCMTS), feasibility-first Pareto GRPO (\FFPGRPO), and adaptive
Pareto-guided policy refinement (\APR). All reported evaluations use one predicted trajectory per scene and require no
test-time scorer, candidate reranking, or model ensemble.

\subsection{Policy-Compatible Multi-Trajectory Supervision}
\label{app:pc-mts-details}

\paragraph{Candidate quality and metric partition.}
The candidate pool combines trajectories generated by historical checkpoints,
structured trajectory expanders, different GRPO seeds, and specialist policies.
PC-MTS is source agnostic: every candidate is evaluated by the same planning
scorer, and the logged trajectory is retained for every scene. Candidate
admission follows three ordered tests. Hard safety and compliance violations are
removed first, a minimum aggregate-quality requirement is then applied, and the
scene-wise Pareto front is finally constructed over the continuous planning
objectives. This order prevents a strong score in one component from compensating
for a hard violation while preserving alternatives that express meaningful
trade-offs.

\begin{table*}[t]
\centering
\small
\setlength{\tabcolsep}{5pt}
\renewcommand{\arraystretch}{1.12}
\caption{Metric roles used throughout AMPT. Feasibility and protected conditions
are checked before Pareto comparison.}
\label{tab:metric-partition}
\begin{tabular}{p{0.10\textwidth}p{0.16\textwidth}p{0.23\textwidth}p{0.29\textwidth}p{0.09\textwidth}}
\toprule
Benchmark & Hard feasibility & Protected conditions & Pareto quality objectives & Aggregate \\
\midrule
NAVSIM v1 & NC, DAC & DDC guard; reference-relative EP and TTC protection & EP, TTC, comfort & PDMS \\
NAVSIM v2 & NC, DAC, DDC, TLC & Reference-relative safety and progress protection & EP, TTC, and quality term $Q$ & EPDMS \\
\bottomrule
\end{tabular}
\end{table*}

For NAVSIM v2, $Q=(\mathrm{LK}+\mathrm{HC}+\mathrm{EC})/3$ is used as a
compact Pareto axis during optimization, while the official evaluation reports
LK, HC, and EC separately. The partition in Table~\ref{tab:metric-partition}
implements the same principle under both protocols: feasibility is
non-compensable, whereas the remaining objectives are compared only inside the
feasible region.

\paragraph{Compatibility with the frozen imitation policy.}
Planning quality alone does not establish that a target is appropriate for the
learner. After GT-only imitation training, the resulting policy is frozen and
used to sample a rollout bank $\mathcal R$ for each scene. Candidate
compatibility is measured by the average normalized waypoint distance to its
$K$ nearest rollouts,
\begin{equation}
 d_{\mathrm{PC}}(\tau,\mathcal R)
 =\frac{1}{K}\sum_{\rho\in\operatorname{KNN}_{K}(\tau,\mathcal R)}
 d(\tau,\rho),
\label{eq:supp-pc-distance}
\end{equation}
where $d$ is computed in the normalized trajectory representation used by the
diffusion planner. The threshold is calibrated on held-out rollouts of the same
frozen policy and is estimated separately for each navigation command. The
criterion therefore adapts to the trajectory scale and multimodality of the
learner instead of imposing a fixed geometric radius around the logged path.

\paragraph{Local feasibility and scene-balanced learning.}
A candidate may be close to the rollout bank yet occupy a narrow feasible
region. PC-MTS therefore generates temporally coherent nearby trajectories at
the scale of the normal variation observed in frozen-policy rollouts. A
candidate is retained only when the resulting local trajectories preserve
feasibility and the protected metrics. Perturbations are applied to complete
trajectory segments rather than independently jittering waypoints, which
preserves temporal structure and avoids physically implausible diagnostic
samples.

During fine-tuning, every scene contributes exactly one supervision target:
either its logged trajectory or one accepted candidate. Consequently, scenes
with more candidates do not receive greater loss weight. Score, Pareto, and
PC-MTS variants share the same candidate-generation pipeline, diffusion
objective, training schedule, and scene-sampling rule; only their admission
criteria differ. PC-MTS therefore changes the trajectory distribution learned
by the policy rather than the amount of optimization assigned to each scene.
Its objective is to obtain an initialization whose sampled trajectories remain
feasible and informative for subsequent GRPO, rather than merely maximizing the
intermediate imitation score.

\subsection{Feasibility-First Pareto GRPO}
\label{app:ff-details}

\paragraph{Coherent scene reference.}
For each scene, the reference is selected from the logged trajectory, the
single-trajectory prediction of the PC-MTS policy, and the retained candidates.
The highest-scoring feasible trajectory is used, so the aggregate score and all
component values come from the same executable plan. This coherent reference
avoids constructing an unattainable target by combining the best EP, TTC, or
comfort values from different trajectories.

\paragraph{Group-adaptive credit assignment.}
Rollouts are divided by feasibility before quality comparison. In NAVSIM v1,
NC and DAC define the hard feasible set, DDC is protected, and a
reference-relative EP floor prevents higher TTC from being obtained merely by
slowing down. Among rollouts that pass these guards, the Pareto front is formed
over EP, TTC, and comfort. NAVSIM v2 applies the same rule with its extended
compliance metrics.

Let $\widehat A_i$ denote the group-normalized aggregate-quality advantage and
let $\mathcal P$ denote the reference-admissible Pareto front. Using
$[x]_+=\max(x,0)$ and $[x]_- = \min(x,0)$, the credit rule is
\begin{equation}
 A_i^{\mathrm{FF}}=
 \begin{cases}
 [\widehat A_i]_+, & \tau_i\in\mathcal P,\\
 [\widehat A_i]_-, & \tau_i\text{ is feasible},\ \tau_i\notin\mathcal P,\\
 -\lambda_{\mathrm{u}}+[\widehat A_i]_-, & \tau_i\text{ is infeasible}.
 \end{cases}
\label{eq:supp-ff-advantage}
\end{equation}
Feasible rollouts that fail a reference guard fall into the second case and
cannot receive positive credit. Aggregate quality determines update strength,
whereas feasibility, reference-level performance, and Pareto comparison decide
whether positive reinforcement is permitted. The distinction is central: an EP
gain cannot offset a hard safety or compliance regression.

\paragraph{Fully infeasible groups.}
When every rollout is infeasible, no sampled trajectory is used as a positive
example. Non-positive advantages rank violation severity, while weak diffusion
supervision from the coherent safe reference gives an explicit direction back
toward the feasible region. These groups are downweighted so repeated failures
do not dominate the batch, and the policy loss retains the KL and behavior
regularization of the underlying GRPO baseline. This branch separates the two
roles that scalar ranking conflates: ranking failures by severity and restoring
probability mass to a known feasible behavior.

\paragraph{Interpretation.}
FF-PGRPO performs scene-local and group-local Pareto comparison; it is not a
global Pareto-policy solver. Its purpose is to prevent an unsafe,
reference-regressing, or dominated rollout from receiving positive credit.
Accordingly, safety recovery and performance improvement are handled by
different learning signals instead of being compressed into a single scalar
ranking.

\subsection{Adaptive Pareto-Guided Policy Refinement}
\label{app:apr-details}

\paragraph{Multi-source teacher pool.}
APR starts from the PC-MTS+FF-PGRPO policy and forms a multi-source trajectory
pool from current-policy trajectories, independent GRPO seeds, structured
expansions, and policies specialized for safety, progress, or trajectory
regularity. The current single-trajectory prediction is the scene baseline. A
candidate is eligible only when it is feasible, improves the aggregate score by
a margin, and does not degrade protected components. Among eligible candidates,
APR favors larger gains with smaller behavioral distance, curvature, and jerk.

\begin{table}[H]
\centering
\small
\setlength{\tabcolsep}{4pt}
\caption{Teacher-pool statistics for one APR construction round. Exact
re-evaluation after interpolation removes endpoint-valid candidates that do not
remain valid local targets.}
\label{tab:teacher-pool-counts}
\begin{tabular}{lrr}
\toprule
Stage & Candidates & Retained \\
\midrule
Teacher source A & 25,402 & 25,402 \\
Teacher source B & 13,859 & 13,859 \\
Merged strict union & 39,261 & 39,261 \\
Post-interpolation audit & 39,261 & 28,910 \\
\bottomrule
\end{tabular}
\end{table}

The post-interpolation audit retains 73.6\% of the strict union and rejects
10,351 candidates. This reduction is substantial: endpoint validity alone does
not guarantee that a target on the segment between the current prediction and
the teacher will remain feasible and high-scoring. Exact re-evaluation is
therefore part of teacher construction rather than a cosmetic post-processing
step.

\paragraph{Local targets and exact re-evaluation.}
A distant teacher is converted into a local target by interpolation,
\begin{equation}
 \widetilde\tau(\alpha)
 =(1-\alpha)\tau_{\mathrm{cur}}+\alpha\tau_{\mathrm{teach}}.
\label{eq:supp-apr-interp}
\end{equation}
A short descending list of interpolation ratios is tested. The first target
that remains feasible, improves the aggregate score, preserves all protected
metrics, and passes the current-policy compatibility test is accepted. The
actual interpolated trajectory is always scored again because valid endpoints
do not imply a valid interior trajectory.

\paragraph{Advantage weighting and retention.}
Accepted targets are weighted by their verified score gain,
\begin{equation}
 w=\operatorname{clip}\!\left(
 \exp\!\left[\frac{R(\widetilde\tau)-R(\tau_{\mathrm{cur}})}{\beta}\right],
 1,w_{\max}\right).
\label{eq:supp-apr-weight}
\end{equation}
Training combines weighted diffusion learning on accepted teachers with
retention on current-policy outputs and a policy-distance regularizer. Scenes
without a valid teacher contribute only to retention. This composition allows
APR to repair residual hard scenes without sacrificing the broad set of
behaviors already handled by the current policy.

After every round, the refined checkpoint becomes the new reference policy. Its
predictions and rollout bank are regenerated, and teacher eligibility is
recomputed. Teachers that have become redundant or regressive are removed,
while candidates that were previously too distant can become admissible after
the policy moves closer to them. Safety-, progress-, and structure-oriented
branches are consolidated with an anchor-constrained task-vector merge, and the
merged checkpoint remains a single policy at inference.

\supplementclearpage

\section{Implementation and Evaluation Protocol}
\label{app:implementation}

\subsection{Architecture and Optimization Scope}

The implementation reuses the InternVL3-2B VLA backbone and the diffusion
planner interface of ReCogDrive. The VLM backbone is kept fixed throughout the
three stages, and no additional trainable module is added to the deployed policy. PC-MTS,
FF-PGRPO, and APR modify the trajectory targets, rollout credit, and policy
updates of the diffusion planner while preserving the perception and language
representation stack. The candidate trajectories are generated from the
training data and training-time policies; no external driving dataset is added.

This design isolates the contribution of data construction and policy
optimization. Improvements in the reported results therefore arise from how
multi-trajectory candidates are selected and transferred, rather than from a
larger backbone, an enlarged training corpus, or a separately adapted VLM.

\subsection{Training Schedule and Hyperparameters}

All four training stages are executed on eight NVIDIA A800 GPUs.

\begin{table*}[t]
\centering
\small
\setlength{\tabcolsep}{4.5pt}
\renewcommand{\arraystretch}{1.12}
\caption{Training schedule. Each stage passes a single policy checkpoint to the
next stage.}
\label{tab:training-schedule}
\begin{tabular}{p{0.11\textwidth}p{0.19\textwidth}p{0.28\textwidth}p{0.14\textwidth}p{0.15\textwidth}}
\toprule
Stage & Initialization & Training signal & Budget & Output \\
\midrule
GT-only IL & Pretrained VLA and diffusion planner & One logged trajectory per scene & 200 epochs & Frozen $\pi_{\mathrm{GT}}$ \\
PC-MTS & Copy of $\pi_{\mathrm{GT}}$ & Logged or accepted policy-compatible trajectory & 20 epochs & $\pi_{\mathrm{initial}}$ \\
FF-PGRPO & $\pi_{\mathrm{initial}}$ & 16 rollouts per scene; Pareto credit and safe-reference recovery & 10 epochs & $\pi_{\mathrm{FF}}$ \\
APR & $\pi_{\mathrm{FF}}$ & Audited local teachers and policy retention & 3 rounds & Final AMPT policy \\
\bottomrule
\end{tabular}
\end{table*}

\begin{table*}[t]
\centering
\small
\setlength{\tabcolsep}{4pt}
\renewcommand{\arraystretch}{1.1}
\caption{Optimization hyperparameters used in the four training stages.}
\label{tab:optimization-config}
\begin{tabular}{lcccc}
\toprule
Setting & GT-only & PC-MTS & FF-PGRPO & APR \\
\midrule
Optimizer & AdamW & AdamW & AdamW & AdamW \\
Peak learning rate & $10^{-4}$ & $10^{-5}$ & $10^{-4}$ & $5\!\times\!10^{-7}$ \\
Minimum learning rate & $10^{-6}$ & $10^{-6}$ & $10^{-6}$ & $2.5\!\times\!10^{-7}$ \\
Weight decay & $10^{-4}$ & $10^{-4}$ & $10^{-4}$ & $10^{-4}$ \\
Batch per GPU & 8 & 8 & 2 & 8 \\
Gradient accumulation & 2 & 2 & 4 & 1 \\
Precision & bf16 & bf16 & bf16 & bf16 \\
Gradient clipping & 1.0 & 1.0 & 1.0 & 1.0 \\
Policy regularization & \NA & \NA & BC $0.10\!\rightarrow\!0.05$; KL 0.02 & Retention 1.0; prior 0.02 \\
\bottomrule
\end{tabular}
\end{table*}

APR uses a score-gain temperature of $0.05$, a maximum teacher weight of $5$,
a jerk penalty of $0.005$, and one selected teacher per scene. The absolute and
reference-relative safety guards use a DDC floor of $0.99$ with a maximum drop
of $0.01$, and a TTC floor of $0.95$ with a maximum drop of $0.02$. These
settings keep teacher selection conservative while allowing measurable progress
improvements.

\subsection{Benchmark and Inference Protocol}

NAVSIM v1 and v2 are evaluated with the official scorers. The v1 evaluation
contains 12,138 valid scored scenes, and the v2 evaluation contains 12,146
scenes. Both protocols use exactly one predicted trajectory per scene. The
rollout bank, group sampling, candidate pool, scorer, and specialist policies
are training-only components.

\begin{table*}[t]
\centering
\small
\setlength{\tabcolsep}{7pt}
\caption{Final single-trajectory result on NAVSIM v1. The main-paper PDMS of
91.4 is the rounded form of 91.45.}
\label{tab:v1-reproduction}
\begin{tabular}{lrrrrrrr}
\toprule
Method & Scenes & NC & DAC & TTC & C & EP & PDMS \\
\midrule
AMPT & 12,138 & 98.53 & 97.99 & 96.00 & 100.00 & 86.75 & 91.45 \\
\bottomrule
\end{tabular}
\end{table*}

\begin{table*}[t]
\centering
\small
\setlength{\tabcolsep}{3.4pt}
\caption{Final single-trajectory result on NAVSIM v2. The main-paper EPDMS of
89.1 is the rounded form of 89.12.}
\label{tab:v2-reproduction}
\begin{tabular}{lrrrrrrrrrrr}
\toprule
Method & Scenes & NC & DAC & DDC & TLC & TTC & EP & LK & HC & EC & EPDMS \\
\midrule
AMPT & 12,146 & 98.42 & 97.69 & 99.00 & 99.71 & 97.66 & 89.45 & 91.73 & 97.09 & 87.74 & 89.12 \\
\bottomrule
\end{tabular}
\end{table*}

\begin{figure*}[t]
\centering
\includegraphics[width=0.82\textwidth]{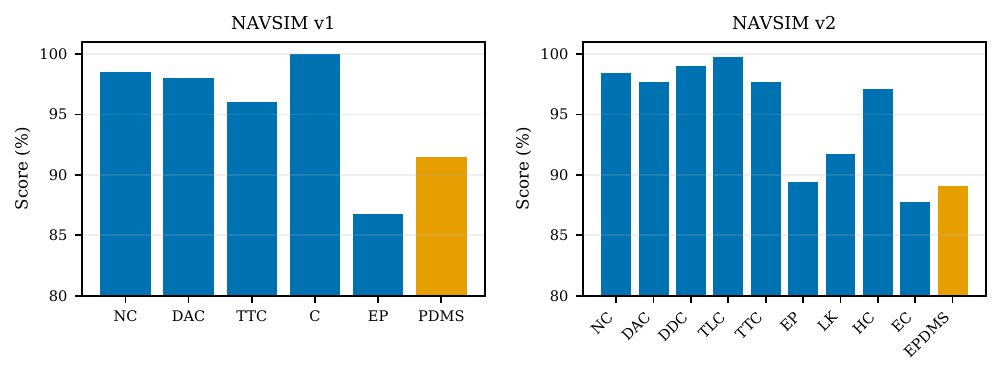}
\caption{Component profiles of the final AMPT policies under NAVSIM v1 and v2.
The two protocols show the same operating pattern: high feasibility and
compliance are maintained while progress is improved inside the feasible
region.}
\label{fig:metric-profiles}
\end{figure*}

The v1 component profile shows that the final score is supported jointly by NC,
DAC, and TTC rather than by EP alone. Under the more detailed v2 protocol, DDC,
TLC, and TTC remain near saturation while EP reaches 89.45. This consistency is
important because v2 introduces additional compliance, lane-keeping, and
comfort terms: the gain is preserved even when the evaluator exposes a richer
set of ways in which aggressive planning can fail.

\subsection{Hard-Scene Recovery Protocol}

The hard subset contains 658 scenes for which the initial ReCogDrive IL policy
produces a collision or drivable-area failure in all five sampling rounds,
resulting in zero PDMS. The recovery comparison uses the scalar-GRPO and AMPT
checkpoints under the same single-trajectory protocol. A scene is counted as
recovered when its evaluated trajectory obtains non-zero PDMS. The paired
transition analysis in Sec.~\ref{app:failure-analysis} reports both recovered
and newly failed scenes.

\subsection{Closest-Method Protocols}

\begin{table}[H]
\centering
\small
\setlength{\tabcolsep}{5pt}
\caption{Inference protocol of the closest compared policies.}
\label{tab:baseline-protocol}
\begin{tabular}{lccc}
\toprule
Method & Traj. & Scorer & Result source \\
\midrule
AMPT & 1 & No & - \\
ReCogDrive & 1 & No & Published \\
DiffusionDriveV2 & 20 & Yes & Published \\
Drive-JEPA & 32 & Yes & Published \\
\bottomrule
\end{tabular}
\end{table}

AMPT preserves the deployment cost of the underlying single-trajectory policy.
All additional computation is confined to training: candidate scoring is used
for supervision construction, group rollouts are used for credit assignment,
and the teacher pool is used for refinement. None of these components changes
the online inference graph.

\supplementfloatbarrier
\supplementclearpage
\supplementonecolumn
\ifdefined\AMPTEmbeddedSupplement
\else
\supplementalgorithmheading
\fi

\supplementalgorithmgroupbegin
\begin{supplementalgorithm}
\ifdefined\AMPTEmbeddedSupplement
\supplementalgorithmheading
\fi
\caption{Policy-Compatible Multi-Trajectory Supervision (PC-MTS)}
\label{alg:pc-mts}
\supplementalgorithmsize
\textbf{Input}: Dataset $\mathcal D=\{(o_s,\tau_s^{\mathrm{gt}})\}$,
candidate pools $\{\mathcal C_s\}$, planning evaluator.\\
\textbf{Parameter}: Quality threshold $\gamma_s$, rollout-bank size
$G_{\mathrm r}$, neighbors $K$, compatibility threshold $\delta_c$,
local-sample count $M_{\mathrm p}$, feasibility ratio $\eta$.\\
\textbf{Output}: Accepted sets $\{\mathcal U_s\}$ and
$\pi_{\mathrm{initial}}$.
\begin{algorithmic}[1]
\STATE Train $\pi_{\mathrm{GT}}$ on logged trajectories with diffusion SFT.
\STATE Freeze $\pi_{\mathrm{GT}}$ and calibrate $\delta_c$ for each command $c$.
\FOR{each scene $s$}
    \STATE $\mathcal Q_s\leftarrow
    \operatorname{PF}(\{\tau\in\mathcal C_s:F(\tau)=1,
    R(\tau)\ge\gamma_s\})$.
    \STATE Sample $\mathcal R_s\sim\pi_{\mathrm{GT}}(\cdot\mid o_s)$,
    $|\mathcal R_s|=G_{\mathrm r}$; set $\mathcal U_s\leftarrow\varnothing$.
    \FOR{each $\tau\in\mathcal Q_s$}
        \IF{$d_{\mathrm{PC}}(\tau,\mathcal R_s)\le\delta_{c_s}$}
            \STATE Generate $M_{\mathrm p}$ temporally coherent nearby trajectories.
            \STATE Let $\kappa(\tau)$ be the fraction that remain feasible.
            \IF{$\kappa(\tau)\ge\eta$}
                \STATE $\mathcal U_s\leftarrow\mathcal U_s\cup\{\tau\}$.
            \ENDIF
        \ENDIF
    \ENDFOR
\ENDFOR
\STATE $\pi_{\mathrm{initial}}\leftarrow\operatorname{Copy}(\pi_{\mathrm{GT}})$.
\FOR{each multi-trajectory SFT update}
    \FOR{each scene $s$ in the minibatch}
        \STATE Sample one target from $\{\tau_s^{\mathrm{gt}}\}\cup\mathcal U_s$.
        \STATE Accumulate its diffusion SFT loss.
    \ENDFOR
    \STATE Update $\pi_{\mathrm{initial}}$.
\ENDFOR
\STATE \textbf{return} $\pi_{\mathrm{initial}},\{\mathcal U_s\}$.
\end{algorithmic}
\end{supplementalgorithm}

\supplementclearpage
\begin{supplementalgorithm}
\caption{Feasibility-First Pareto GRPO (FF-PGRPO)}
\label{alg:ff-pgrpo}
\supplementalgorithmsize
\textbf{Input}: $\pi_{\mathrm{initial}}$, dataset $\mathcal D$,
accepted sets $\{\mathcal U_s\}$, planning evaluator.\\
\textbf{Parameter}: Group size $G$, unsafe penalty $\lambda_{\mathrm u}$,
recovery weight $\lambda_{\mathrm{rec}}$, policy regularization $\beta$.\\
\textbf{Output}: Pareto-optimized policy $\pi_{\mathrm{FF}}$.
\begin{algorithmic}[1]
\FOR{each scene $s$}
    \STATE $\mathcal H_s\leftarrow
    \{\tau_s^{\mathrm{gt}},\operatorname{Deploy}(\pi_{\mathrm{initial}},o_s)\}
    \cup\mathcal U_s$.
    \STATE Set $\tau_s^{\mathrm{ref}}$ to the highest-scoring feasible
    trajectory in $\mathcal H_s$.
\ENDFOR
\STATE $\pi_{\mathrm{FF}}\leftarrow\operatorname{Copy}(\pi_{\mathrm{initial}})$.
\FOR{each GRPO update}
    \FOR{each scene $s$ in the minibatch}
        \STATE Sample $\{\tau_i\}_{i=1}^{G}\sim
        \pi_{\mathrm{FF}}(\cdot\mid o_s)$ and evaluate all metrics.
        \STATE Compute group-normalized quality advantages $\{\widehat A_i\}$.
        \IF{$\sum_iF(\tau_i)>0$}
            \STATE $\mathcal V_s\leftarrow
            \{\tau_i:F(\tau_i)=1,
            \operatorname{Guard}(\tau_i,\tau_s^{\mathrm{ref}})=1\}$.
            \STATE $\mathcal P_s\leftarrow\operatorname{PF}(\mathcal V_s)$.
            \FOR{$i=1,\ldots,G$}
                \IF{$\tau_i\in\mathcal P_s$}
                    \STATE $A_i\leftarrow\max(\widehat A_i,0)$.
                \ELSIF{$F(\tau_i)=1$}
                    \STATE $A_i\leftarrow\min(\widehat A_i,0)$.
                \ELSE
                    \STATE $A_i\leftarrow-
                    \lambda_{\mathrm u}+\min(\widehat A_i,0)$.
                \ENDIF
            \ENDFOR
            \STATE $\mathcal L_{\mathrm{rec}}\leftarrow0$.
        \ELSE
            \STATE Assign violation-aware advantages $A_i\le0$.
            \STATE $\mathcal L_{\mathrm{rec}}\leftarrow
            \mathcal L_{\mathrm{SFT}}(o_s,\tau_s^{\mathrm{ref}})$.
        \ENDIF
        \STATE Accumulate $\mathcal L_{\mathrm{GRPO}}(\{A_i\})+
        \lambda_{\mathrm{rec}}\mathcal L_{\mathrm{rec}}+
        \beta\mathcal L_{\mathrm{reg}}$.
    \ENDFOR
    \STATE Update $\pi_{\mathrm{FF}}$.
\ENDFOR
\STATE \textbf{return} $\pi_{\mathrm{FF}}$.
\end{algorithmic}
\end{supplementalgorithm}
\supplementalgorithmgroupend
\supplementfloatbarrier

\supplementclearpage
\supplementalgorithmthree{%
\caption{Adaptive Pareto-Guided Policy Refinement (APR)}
\label{alg:apr}
\supplementalgorithmsize
\textbf{Input}: $\pi_{\mathrm{FF}}$, dataset $\mathcal D$,
multi-source pools $\{\mathcal C_s\}$, planning evaluator, development set.\\
\textbf{Parameter}: Maximum rounds $K_{\mathrm{APR}}$, score margin
$\delta_R$, interpolation list $\mathcal A$, temperature $\beta_w$,
maximum weight $w_{\max}$, retention weight $\lambda_{\mathrm{ret}}$.\\
\textbf{Output}: Refined policy $\pi^\star$.
\begin{algorithmic}[1]
\STATE $\pi^{(0)}\leftarrow\pi_{\mathrm{FF}}$; $k\leftarrow0$.
\WHILE{$k<K_{\mathrm{APR}}$}
    \STATE $\mathcal T^{(k)}\leftarrow\varnothing$.
    \FOR{each scene $s$}
        \STATE $\tau_s^{\mathrm{cur}}\leftarrow
        \operatorname{Deploy}(\pi^{(k)},o_s)$.
        \STATE Sample rollout bank $\mathcal R_s^{(k)}$ and recalibrate
        compatibility.
        \STATE $\mathcal B_s\leftarrow\{\tau\in\mathcal C_s:
        \operatorname{Audit}(\tau,\tau_s^{\mathrm{cur}})=1\}$.
        \STATE Rank $\mathcal B_s$ by score gain, distance, curvature, and jerk.
        \STATE $\mathrm{accepted}\leftarrow\mathrm{false}$.
        \FOR{each ranked teacher $\tau^{\mathrm{teach}}$}
            \FOR{each $\alpha\in\mathcal A$ in descending order}
                \STATE $\widetilde\tau\leftarrow
                (1-\alpha)\tau_s^{\mathrm{cur}}+\alpha\tau^{\mathrm{teach}}$.
                \STATE Re-evaluate $\widetilde\tau$.
                \IF{$\operatorname{Audit}(\widetilde\tau,
                    \tau_s^{\mathrm{cur}})=1$ and
                    $d_{\mathrm{PC}}(\widetilde\tau,\mathcal R_s^{(k)})
                    \le\delta_{c_s}^{(k)}$}
                    \STATE $w\leftarrow\operatorname{clip}\!\left(
                    \exp\!\left[\frac{R(\widetilde\tau)-
                    R(\tau_s^{\mathrm{cur}})}{\beta_w}\right],1,w_{\max}\right)$.
                    \STATE $\mathcal T^{(k)}\leftarrow
                    \mathcal T^{(k)}\cup\{(s,\widetilde\tau,w)\}$.
                    \STATE $\mathrm{accepted}\leftarrow\mathrm{true}$;
                    \textbf{break}.
                \ENDIF
            \ENDFOR
            \IF{$\mathrm{accepted}$}
                \STATE \textbf{break}.
            \ENDIF
        \ENDFOR
    \ENDFOR
    \STATE $\widetilde\pi\leftarrow\operatorname{Copy}(\pi^{(k)})$.
    \STATE Train $\widetilde\pi$ with weighted teacher SFT, policy retention,
    and policy-distance regularization.
    \STATE Consolidate specialist branches around $\pi^{(k)}$ with the
    anchor-constrained task-vector merge.
    \IF{$\widetilde\pi$ improves the development metric}
        \STATE $\pi^{(k+1)}\leftarrow\widetilde\pi$; $k\leftarrow k+1$.
        \STATE Rebuild current-policy predictions and teacher sets.
    \ELSE
        \STATE \textbf{break}.
    \ENDIF
\ENDWHILE
\STATE $\pi^\star\leftarrow\pi^{(k)}$.
\STATE \textbf{return} $\pi^\star$.
\end{algorithmic}
}

\supplementclearpage
\supplementtwocolumn

\section{Analysis of Policy-Compatible Supervision}
\label{app:pc-analysis}

\subsection{The Supervision--Optimization Reversal}

\begin{table}[H]
\centering
\small
\setlength{\tabcolsep}{7pt}
\caption{Effect of trajectory-set construction. Every imitation checkpoint is
followed by the same original scalar GRPO.}
\label{tab:supervision-reversal}
\begin{tabular}{lrrr}
\toprule
Supervision & IL PDMS & Post-GRPO & Change \\
\midrule
GT-only & 86.4 & 90.3 & $+3.9$ \\
Score & 87.2 & 85.8 & $-1.4$ \\
Pareto & \textbf{87.5} & 86.7 & $-0.8$ \\
PC-MTS & 86.9 & \textbf{91.1} & \textbf{$+4.2$} \\
\bottomrule
\end{tabular}
\end{table}

\begin{figure}[H]
\centering
\includegraphics[width=\columnwidth]{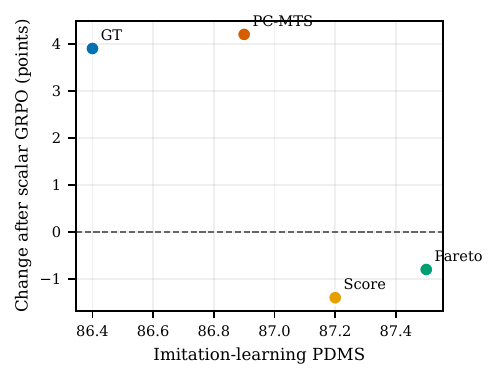}
\caption{Imitation performance and the change produced by scalar GRPO. Higher
imitation performance does not predict a better optimization outcome.}
\label{fig:supervision-mismatch}
\end{figure}

Table~\ref{tab:supervision-reversal} exposes a ranking reversal between
imitation and post-training. Score and Pareto improve the imitation checkpoint
by 0.8 and 1.1 points over GT-only training, yet the same scalar GRPO reduces
their scores by 1.4 and 0.8 points. PC-MTS obtains a smaller direct imitation
gain but receives a 4.2-point improvement from GRPO. Relative to Score and
Pareto, the difference in GRPO response is 5.6 and 5.0 points, respectively.
The magnitude of this reversal is considerably larger than the differences
between the imitation checkpoints themselves.

The result separates trajectory quality from optimization utility. Aggregate
and component-wise Pareto quality identify desirable plans, but they do not
show whether fitting those plans preserves probability mass over feasible
rollouts. PC-MTS changes the learned sampling distribution before GRPO: its
benefit is therefore expressed primarily in the policy's response to
post-training. The one-target-per-scene rule is important here because it
prevents a larger candidate set from mechanically increasing a scene's gradient
weight. The observed difference follows from which behaviors are learned, not
from how many losses a scene contributes.

\subsection{Why Policy Compatibility Differs from GT Proximity}

Candidate-to-GT distance and policy compatibility answer different questions.
A diffusion policy can represent a multimodal neighborhood around the logged
trajectory. A candidate may be moderately distant from GT yet lie close to one
of the policy's sampled modes; conversely, a geometrically close candidate may
lie in a direction to which the learned policy assigns little probability.
Using the rollout bank as the reference therefore measures compatibility with
the learner that will absorb the target, rather than similarity to one recorded
future.

This policy-relative view also explains the relationship between PC-MTS and
APR. PC-MTS favors local extensions that can be learned without destabilizing
the initial policy. APR later recalibrates compatibility around the improved
policy, allowing candidates that were previously too distant to enter the
teacher set. The two stages form a curriculum whose boundary moves with policy
capability.

\supplementclearpage
\section{Analysis of Feasibility-First Pareto GRPO}
\label{app:ff-analysis}

\subsection{Complementary Roles of the Three Stages}

\begin{table*}[t]
\centering
\small
\setlength{\tabcolsep}{5.5pt}
\caption{Ablation of PC-MTS, FF-PGRPO, and APR on NAVSIM v1.}
\label{tab:stage-ablation}
\begin{tabular}{lcccrrrrrr}
\toprule
Configuration & PC-MTS & FF-PGRPO & APR & NC & DAC & TTC & C & EP & PDMS \\
\midrule
GT-only &  &  &  & 98.1 & 94.7 & 94.2 & 100.0 & 80.9 & 86.5 \\
PC-MTS & $\checkmark$ &  &  & 98.2 & 95.2 & 94.5 & 100.0 & 81.3 & 86.9 \\
FF-PGRPO only &  & $\checkmark$ &  & 98.4 & 98.0 & 95.8 & 100.0 & 85.93 & 91.0 \\
PC-MTS + FF-PGRPO & $\checkmark$ & $\checkmark$ &  & 98.5 & 98.0 & 96.0 & 100.0 & 86.0 & 91.1 \\
Full AMPT & $\checkmark$ & $\checkmark$ & $\checkmark$ & 98.5 & 98.0 & 96.0 & 100.0 & 86.7 & \textbf{91.4} \\
\bottomrule
\end{tabular}
\end{table*}

FF-PGRPO provides the largest direct policy gain. Relative to the GT-only row,
it increases DAC by 3.3 points, TTC by 1.6 points, and EP by 5.03 points,
raising PDMS from 86.5 to 91.0. The simultaneous improvement of DAC, TTC, and
EP is the central result: efficiency is not obtained by relaxing feasibility or
reducing the safety margin.

PC-MTS adds only 0.1 PDMS when placed before FF-PGRPO, but this aggregate
difference understates its role. The selection study shows that conventional
multi-trajectory supervision can reverse the benefit of GRPO altogether. PC-MTS
therefore serves as a distributional safeguard for the policy-optimization
stage, while FF-PGRPO provides the main on-policy improvement. APR then raises
EP from 86.0 to 86.7 without changing the displayed NC, DAC, TTC, or comfort,
recovering residual efficiency after the safety-sensitive metrics have
stabilized.

\subsection{What the Credit Rule Changes}

A direct diagnostic of credit semantics is the positive-credit violation rate,
\begin{equation}
 \mathrm{PCVR}=\frac{N^{+}_{\mathrm{viol}}}{\max(N^{+},1)},
\label{eq:pcvr}
\end{equation}
where $N^{+}$ is the number of positive-advantage rollouts and
$N^{+}_{\mathrm{viol}}$ counts those violating a protected condition. Scalar
GRPO can assign positive credit to such a rollout whenever its aggregate score
is above the group mean. FF-PGRPO enforces $\mathrm{PCVR}=0$ for the encoded
feasibility and reference conditions by construction. This property is stronger
than simply adding another penalty term: protected violations are excluded from
positive reinforcement rather than traded against progress on a common scale.

The all-infeasible branch addresses a different failure of scalar ranking. A
relative score can identify which sampled failure is less severe, but it cannot
place a target inside the feasible region. Safe-reference supervision provides
that missing direction. Downweighting the group then prevents a small number of
hard scenes from dominating updates for the broader dataset. Together, the
negative violation signal and positive recovery target turn fully infeasible
groups from uninformative batches into structured recovery examples.

\supplementclearpage
\section{Analysis of Adaptive Policy Refinement}
\label{app:apr-analysis}

\subsection{Round-Wise Progression}

\begin{table}[H]
\centering
\small
\setlength{\tabcolsep}{8pt}
\caption{APR progression. Round 0 is the policy before refinement.}
\label{tab:apr-rounds}
\begin{tabular}{lrr}
\toprule
Policy & PDMS & Gain \\
\midrule
Round 0 & 91.10 & \NA \\
Round 1 & 91.21 & $+0.11$ \\
Round 2 & 91.37 & $+0.16$ \\
Round 3 & 91.45 & $+0.08$ \\
\bottomrule
\end{tabular}
\end{table}

\begin{figure}[H]
\centering
\includegraphics[width=\columnwidth]{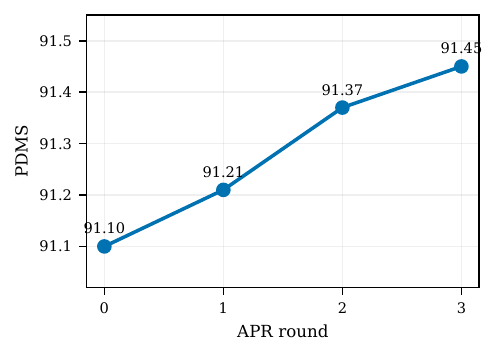}
\caption{PDMS across APR rounds.}
\label{fig:apr-rounds}
\end{figure}

APR starts from a policy that has already captured most of the online gain, so
its role is residual refinement rather than broad recovery. The three rounds
increase PDMS monotonically from 91.10 to 91.45. The final 0.35-point gain is
concentrated in EP, while the displayed NC, DAC, TTC, and comfort values remain
unchanged. This profile matches the teacher audit: APR accepts only candidates
that improve the current policy without crossing the protected metric guards.

The distribution of per-round gains is also informative. The largest increment
appears in Round 2 rather than Round 1, indicating that teacher reconstruction
is not equivalent to repeatedly fitting a fixed set. The first update changes
the policy neighborhood; the next audit can then expose improvements that were
previously too distant or insufficiently advantageous. The final round yields a
smaller increment as the remaining teacher pool approaches the current policy's
performance frontier.

\subsection{Teacher Audit and Dynamic Validity}

The teacher-pool audit in Table~\ref{tab:teacher-pool-counts} rejects 26.4\% of
candidates after interpolation and exact re-evaluation. These candidates satisfy
the endpoint constraints but fail as actual local targets. This distinction
justifies both interpolation and post-interpolation scoring: a teacher is useful
only if the trajectory that the learner is asked to imitate is itself feasible
and improving.

Teacher validity is recomputed after each round because it is defined relative
to the current policy. A teacher can expire when its score advantage disappears,
and a previously distant candidate can become admissible after the policy moves
closer to it. Dynamic reconstruction therefore prevents stale supervision and
turns the multi-source pool into an adaptive curriculum rather than a static
pseudo-label set.

\supplementclearpage
\section{Failure Recovery and Qualitative Analysis}
\label{app:failure-analysis}

\subsection{Gross and Net Recovery}

On the fixed 658-scene hard subset, scalar GRPO produces 367 positive and 291
zero outcomes, whereas AMPT produces 440 positive and 218 zero outcomes. The
recovery rate increases from 55.8\% to 66.9\%, an absolute gain of 11.1
percentage points.

\begin{table}[H]
\centering
\small
\setlength{\tabcolsep}{5pt}
\caption{Recovery on the fixed 658-scene hard subset. Intervals are Wilson 95\%
confidence intervals over scenes.}
\label{tab:failure-recovery}
\begin{tabular}{lrrc}
\toprule
Method & Positive & Zero & Recovery rate \\
\midrule
Scalar GRPO & 367 & 291 & 55.8\% $[52.0,59.5]$ \\
AMPT & 440 & 218 & 66.9\% $[63.2,70.4]$ \\
\bottomrule
\end{tabular}
\end{table}

\begin{table}[H]
\centering
\small
\setlength{\tabcolsep}{7pt}
\caption{Paired state transitions on the same 658 scenes.}
\label{tab:failure-transition}
\begin{tabular}{lrrr}
\toprule
Scalar GRPO & AMPT zero & AMPT positive & Total \\
\midrule
Zero & 198 & 93 & 291 \\
Positive & 20 & 347 & 367 \\
\bottomrule
\end{tabular}
\end{table}

The paired matrix separates gross recovery from net improvement. AMPT repairs
93 scenes that scalar GRPO leaves at zero and introduces 20 regressions, giving
a net repair of 73 scenes. Thus, 78.5\% of the gross repairs remain after
accounting for regressions. This net result is exactly the additional recovery
reported in the main paper and shows that the gain is not produced by merely
trading one set of failures for another.

\begin{figure*}[t]
\centering
\begin{minipage}[t]{0.48\textwidth}
\centering
\includegraphics[width=\linewidth]{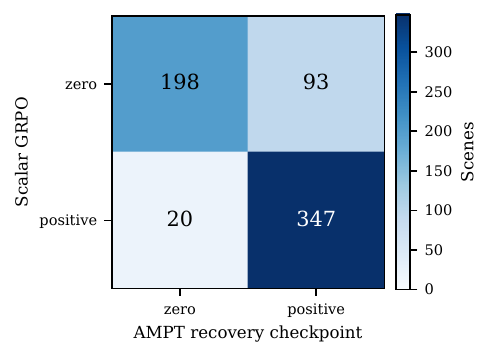}\\[-2pt]
\small (a) Paired state transitions.
\end{minipage}
\hfill
\begin{minipage}[t]{0.48\textwidth}
\centering
\includegraphics[width=\linewidth]{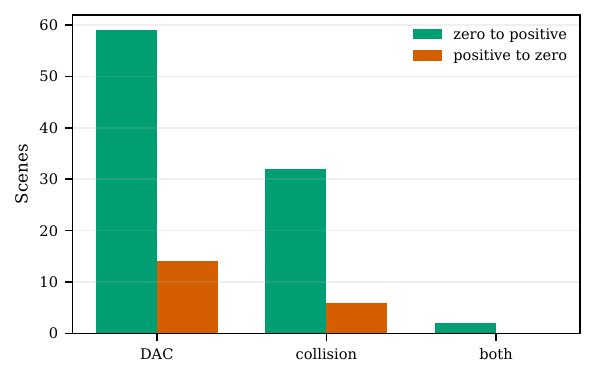}\\[-2pt]
\small (b) Repairs and regressions by initial failure type.
\end{minipage}
\caption{Failure-recovery analysis on the fixed hard subset. The transition
view separates repairs from regressions, and the cause breakdown identifies
where the net recovery is obtained.}
\label{fig:failure-analysis}
\end{figure*}

\subsection{Failure-Type Breakdown}

\begin{table}[H]
\centering
\small
\setlength{\tabcolsep}{5pt}
\caption{Net repair by initial failure type.}
\label{tab:failure-causes}
\begin{tabular}{lrrrr}
\toprule
Failure type & Scenes & Repairs & Regressions & Net \\
\midrule
DAC-only & 480 & 59 & 14 & 45 \\
Collision-only & 165 & 32 & 6 & 26 \\
Collision + DAC & 13 & 2 & 0 & 2 \\
\bottomrule
\end{tabular}
\end{table}

DAC-only failures contribute 45 of the 73 net repairs (61.6\%), while
collision-only scenes contribute 26 (35.6\%). The corresponding net-repair
rates are 9.4\% and 15.8\%, showing that the improvement is not restricted to
the more common drivable-area failures. The mixed subset contributes the
remaining two repairs. This decomposition is consistent with FF-PGRPO: its
feasibility gate directly protects both collision and drivable-area compliance,
while safe-reference recovery supplies a feasible target when the sampled group
contains neither.

The 20 regressions clarify the role of policy retention in APR. A policy can
produce a strong net recovery while still damaging a minority of scenes that
were previously positive. Retention and checkpoint promotion are therefore
necessary complements to targeted teacher distillation; they preserve the broad
successful distribution while difficult scenes receive additional supervision.

\subsection{Representative Transitions}

\begin{itemize}
    \item \textbf{Full repair: \path{00fcad6d092c5e8e} (left).}
    PDMS changes from 0 to 1.000 as NC and TTC both change from 0 to 1,
    yielding complete component recovery.

    \item \textbf{Drivable-area repair: \path{03aa8a0576a25b63} (right).}
    PDMS changes from 0 to 0.963. DAC changes from 0 to 1 while NC and TTC
    remain feasible, showing that recovery does not require a more aggressive
    safety profile.

    \item \textbf{Partial repair: \path{1148c72f141c532d} (left).}
    NC and DAC change from 0 to 1, while TTC remains 0. The resulting PDMS of
    0.583 illustrates that restoring hard feasibility does not imply that every
    continuous component is solved.

    \item \textbf{Persistent failure: \path{00016f8b45c25a1d} (left).}
    PDMS remains 0 because DAC and EP remain 0. This case illustrates the
    residual ceiling imposed by candidate coverage and local policy capability.

    \item \textbf{Regression: \path{4b4a268bee4c5ab5} (left).}
    PDMS changes from 1 to 0 after DAC changes from 1 to 0. The case motivates
    retention and checkpoint-promotion checks and is included to avoid
    survivorship bias.
\end{itemize}

The repaired examples distinguish complete recovery from partial recovery:
AMPT can restore all failed components, or it can first re-enter the feasible
region while a continuous objective remains weak. The persistent and regressed
cases complement the successful examples by showing that candidate quality,
policy locality, and behavior retention jointly determine the final outcome.

\begin{figure*}[t]
\centering
\includegraphics[width=0.92\textwidth]{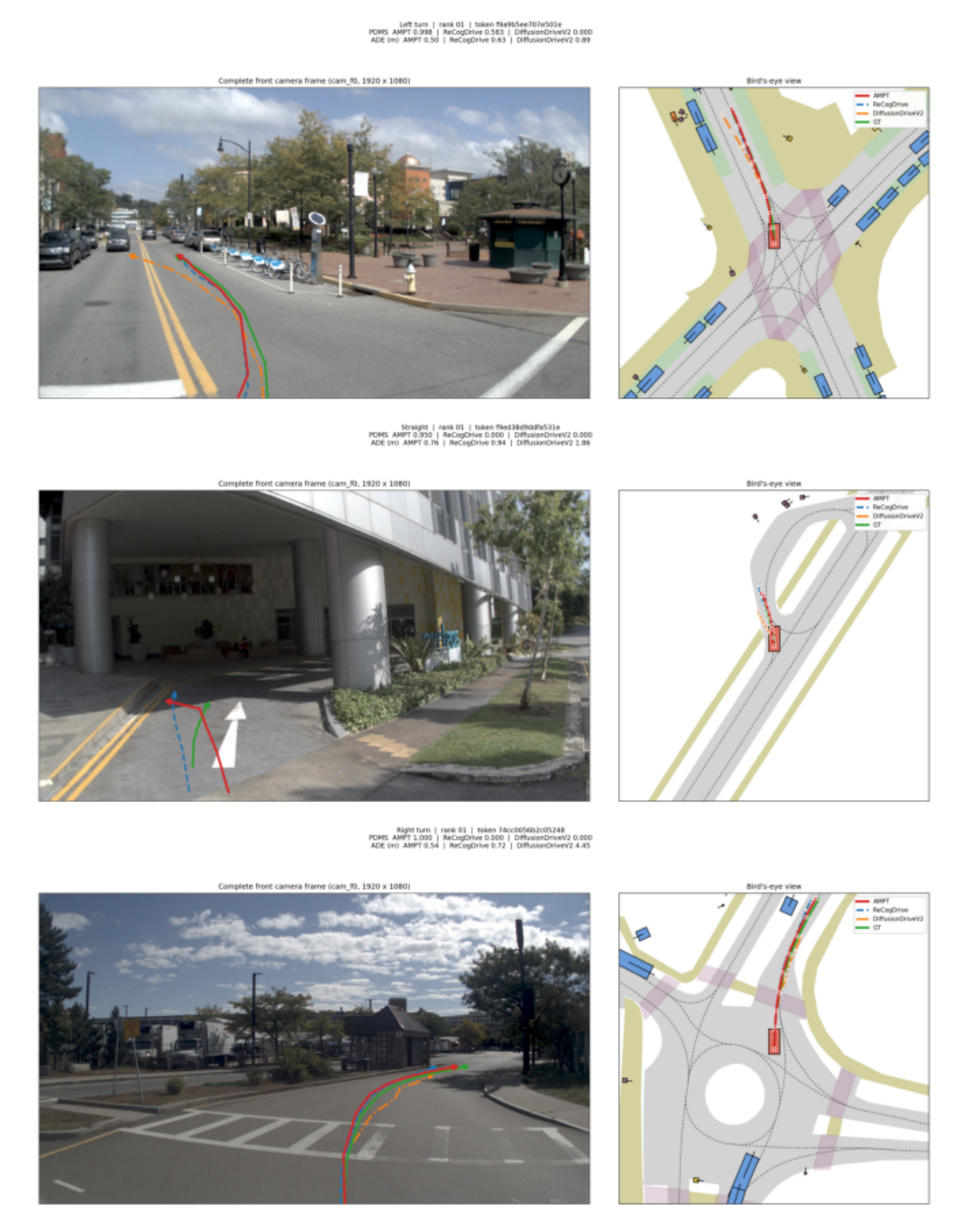}
\caption{Selected single-trajectory predictions from the final AMPT policy for
left, straight, and right commands.}
\label{fig:qualitative-cases}
\end{figure*}

\supplementclearpage
\section{Discussion and Limitations}
\label{app:limitations}

\subsection{Orthogonality to Representation Learning}

AMPT operates after the visual and language representations have been defined.
It neither changes the VLM architecture nor introduces an auxiliary feature
encoder, cross-modal fusion module, representation objective, or semantic
alignment loss. Its intervention is concentrated at the trajectory level:
constructing the supervision set, assigning rollout credit, and validating
policy-relative teachers. Consequently, advances in feature extraction,
representation learning, multimodal fusion, or semantic alignment can be
combined with AMPT by replacing the shared VLA backbone while retaining the
same trajectory-selection and policy-optimization pipeline.

This orthogonality also sharpens the interpretation of the experiments. Because
the backbone, training data, and parameter count remain fixed, the reported
improvements isolate the value of coordinating multi-trajectory supervision
with downstream optimization. AMPT is therefore complementary to model-scaling
and representation-centric research rather than a competing replacement for
those directions.

\subsection{Relation to Conservative Policy Improvement}

PC-MTS shares the support-aware motivation of conservative policy improvement:
BEAR constrains a policy toward the behavior distribution, SPIBB falls back to
a baseline in uncertain regions, and AWAC weights behavior learning by
estimated advantage~\cite{kumar2019bear,laroche2019spibb,nair2020awac}. AMPT
uses related principles at the trajectory level but addresses a different
interface. Compatibility is applied before imitation fine-tuning, Pareto credit
is restricted to the feasible region, and teacher validity is recomputed as the
policy changes. CLOVER is the closest evaluator-guided trajectory-refinement
method, but it retains candidate scoring at inference, whereas AMPT produces one
trajectory without a test-time scorer or reranking~\cite{ang2026clover}.

\subsection{Limitations}

\paragraph{Finite policy samples.}
Compatibility is estimated from a finite rollout bank. Sparse navigation
commands or highly multimodal scenes increase the sampling cost required to
represent the local trajectory distribution accurately.

\paragraph{Evaluator dependence.}
All three stages use the NAVSIM evaluator for filtering, credit assignment, or
teacher validation. Evaluator bias can therefore propagate through the
pipeline, and non-reactive simulation cannot establish a closed-loop safety
guarantee.

\paragraph{Candidate-pool ceiling.}
APR can select only behaviors represented in its multi-source pool. If none of
the sources proposes a feasible improvement, teacher reconstruction cannot
repair the scene.

\paragraph{Locality versus long-horizon value.}
PC-MTS intentionally favors learnable local targets. A distant behavior that
requires a longer curriculum may be rejected at the initial stage, although
dynamic teacher reconstruction can admit it after the policy has improved.

\paragraph{Benchmark-specific hierarchy.}
The feasibility gates, protected metrics, and Pareto objectives follow NAVSIM.
Applying the framework to another benchmark requires defining the corresponding
metric hierarchy and calibration thresholds.

\paragraph{Parameter merging.}
Anchor-constrained task-vector merging limits parameter displacement and
consolidates complementary specialist branches, but behavioral validity remains
determined by the trajectory evaluator and checkpoint-promotion protocol.

\subsection{Reproducibility Assets}

The supplementary package contains the candidate and scene manifests, training
configurations, rollout calibration protocol, FF-PGRPO group and recovery
settings, APR teacher and retention settings, scene-level NAVSIM v1/v2 exports,
and the paired hard-scene transition data. These assets support direct
reproduction of the tables, figures, and analyses in this document.

\supplementbibliography

\supplementend
\endgroup


\end{document}